\documentclass[twoside,11pt]{article}

\usepackage{jmlr2e}
\usepackage{algorithm}
\usepackage{adjustbox}
\usepackage{algpseudocode}
\usepackage{amssymb}
\usepackage{tikz}
\usepackage{verbatim}
\usepackage{subfig}
\usepackage{booktabs}
\usepackage{multirow}
\usepackage{siunitx}
\usepackage[T1]{fontenc}
\usepackage{beramono}
\usepackage{amsthm}

\theoremstyle{definition}
\newtheorem{ex}{Example}[section]

\usepackage[utf8]{inputenc}
\usepackage[english]{babel}

\usepackage{amsthm}

\theoremstyle{definition}
\newtheorem{definition}{Definition}[section]

\usepackage[disable]{todonotes}
\usepackage{todonotes}
\newcommand{\liz}[2][]{\todo[color=cyan!80, #1, inline]{Liz: {\small #2}}}
\newcommand{\ls}[2][]{\todo[color=magenta!80, #1, inline]{Liz: {\small #2}}}
\newcommand{\tm}[2][]{\todo[color=green!80, #1,inline]{Tim: {\small #2}}}

\newcommand{\pmm}[2][]{\todo[color=yellow!80, #1,inline]{Prashan: {\small #2}}}

\ShortHeadings{Distal Explanations for Model-free Explainable Reinforcement Learning}{Prashan Madumal, Tim Miller, Liz Sonenberg and Frank Vetere}
\firstpageno{1}

\begin{document}

\title{Distal Explanations for Model-free Explainable Reinforcement Learning}
\author{}

\editor{}

\maketitle


\tm{I know I said to extend the abstract and intro, but the abstract is perhaps a bit too long now. I suggest restructuring a few ways: (1) start out with a sentence that is ``In this paper, we..." to drive home the topic immediately; (2) then get straight into the first user study, but give a bit less details; (3) then mention the finding of opportunity chains and explain these; and (4) we show how to generate these, calling them `distal explanations', and a brief overview of the study/results; but cut back on some of these details.}
\pmm{have cut back some and summarised}
\ls{have further tightened the abstract}

\begin{abstract}
In this paper we introduce and evaluate a \emph{distal explanation} model for model-free reinforcement learning agents that can generate explanations for `why' and `why not' questions. Our starting point is the observation that causal models can generate \emph{opportunity chains} that take the form of `A enables B and B causes C'. Using insights from  
an analysis of 240 explanations generated in a human-agent experiment, we define a distal explanation model that can analyse counterfactuals and opportunity chains using decision trees and causal models. A recurrent neural network is employed to learn opportunity chains, and decision trees are used to improve the accuracy of task prediction and the generated counterfactuals. We computationally evaluate the model in 6 reinforcement learning benchmarks using different reinforcement learning algorithms. From  a study with 90 human participants, we show that our distal explanation model results in improved outcomes over three scenarios compared with two baseline explanation models.
\end{abstract}



\section{Introduction}

Understanding how artificially intelligent systems behave and make decisions has long since been a topic of interest, and in recent years has resurfaced as `Explainable AI' (XAI). The ability to provide explanations of the behaviour of these systems is particularly important in scenarios where humans need to collaborate with intelligent agents. Often, the success of these collaborative tasks depends on how well the human understands both the long-term goals and immediate actions of the agent.

Explanation models that emulate human models of explanations have the potential to provide intuitive and natural explanations, allowing the human a deeper understanding of the agent~\citep{de2017,abdul2018trends,miller2018explanation,wang2019designing}. There exists a large body of literature in cognitive psychology that studies the  nature of explanations. One prevalent theory is that explanations are innately \emph{causal}~\citep{halpern2005causes}. Causal explanations resonate with humans as we make use of \emph{causal} models of the world to encode cause-effect relationships in our mind~\citep{sloman2005causal}, and leverage these models to explain why \emph{events} happen. Causal models also enable the generation of \emph{counterfactual} explanations---explanations about events that did not happen but could have under different circumstances~\citep{halpern2005causes}. So causal explanations have the potential to provide `better' explanations to humans.

Recent work in the XAI research community has demonstrated the effectiveness of causality and causal explanations for interpretability and explainability~\citep{byrne2019counterfactuals,klein2018explaining,gunning2019darpa,schwab2019cxplain,madumal2019explainable}. In the context of model-free reinforcement learning (RL) agents, causal models have been encoded using \emph{action influence graphs} to generate explanations using \emph{causal chains}~\citep{madumal2019explainable} and have been shown to support subjectively `better' explanations and yield improved performance in \emph{task prediction}~\citep{hoffman2018metrics} as compared with state-action based explanations~\citep{khan2009minimal}. While action influence models provide a skeleton to generate causal explanations for RL agents, finer details of the composition of causal explanations can be absent. We argue that, through investigating interactions of RL agents and humans, some shortcomings of action influence models can potentially be alleviated.  

\tm{A recent review that Abeer got made a good point about this type of approach not really be 'grounded theory' because it is not iterative. It is more 'thematic analysis'; see `Using thematic analysis in psychology'
 Braun and  Clarke. It doesn't invalide the results, but it does mean those researchers in psych/design etc., wouldn't go nuts over it :)}
 \pmm{changed the wording to themtic analysis}

To ground the effect that explanation models have on human explanation, we conduct human-agent
experiments on how humans formulate explanations of agent behaviour. Participants of this study received explanations from three different models: visual agent behaviour explanations; state-action based explanations~\citep{khan2009minimal}; and causal explanations~\citep{madumal2019explainable}. Then participants were asked to formulate their own explanations of the agents' behaviour as a textual input. The study was carried out with 30 participants and we obtained 240 explanations in total. We used thematic analysis~\citep{braun2006using} to identify recurring concepts present in the explanations. 

Results of our analysis show that while causality was indeed present, these self-provided explanations predominantly referred to a \emph{future} action that was dependent on the current action. Participants' tendency to include a future action in their explanations indicates an understanding of the causal chain of actions and events. This phenomenon is well explored in cognitive psychology and is defined as \emph{opportunity chains}~\citep{hilton2005course}. We use insights gained from the human-agent study to inform our design of an explanation model that can explain opportunity chains and the future action termed the \emph{distal} action.
\ls{"Further support our hypothesis" - isn't clear as it isn't explicit what the hypothesis is here, also hard to interpret what a 2010 paper you haven't discussed yet might be saying that is relevant; especially when here you say it's similar, later you explain differences; maybe the solution is to put the claim about 'support' in the discussion rather than the intro?}
\pmm{removed this from intro}

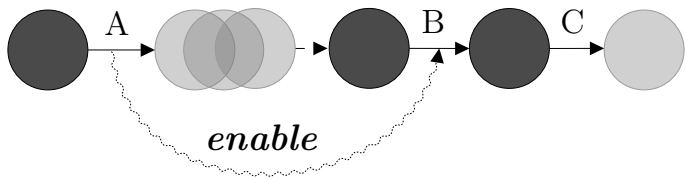
\begin{figure}[!ht]
        \centering
        \begin{adjustbox}{width=90mm,center}
\begin{tikzpicture}[x=0.75pt,y=0.75pt,yscale=-1,xscale=1]

\draw  [color={rgb, 255:red, 0; green, 0; blue, 0 }  ,draw opacity=1 ][fill={rgb, 255:red, 74; green, 74; blue, 74 }  ,fill opacity=1 ] (59.8,125.6) .. controls (59.8,111.79) and (70.99,100.6) .. (84.8,100.6) .. controls (98.61,100.6) and (109.8,111.79) .. (109.8,125.6) .. controls (109.8,139.41) and (98.61,150.6) .. (84.8,150.6) .. controls (70.99,150.6) and (59.8,139.41) .. (59.8,125.6) -- cycle ;
\draw  [color={rgb, 255:red, 0; green, 0; blue, 0 }  ,draw opacity=1 ][fill={rgb, 255:red, 74; green, 74; blue, 74 }  ,fill opacity=1 ] (260.8,124.6) .. controls (260.8,110.79) and (271.99,99.6) .. (285.8,99.6) .. controls (299.61,99.6) and (310.8,110.79) .. (310.8,124.6) .. controls (310.8,138.41) and (299.61,149.6) .. (285.8,149.6) .. controls (271.99,149.6) and (260.8,138.41) .. (260.8,124.6) -- cycle ;
\draw  [color={rgb, 255:red, 0; green, 0; blue, 0 }  ,draw opacity=1 ][fill={rgb, 255:red, 74; green, 74; blue, 74 }  ,fill opacity=1 ] (348.5,124.67) .. controls (348.5,110.86) and (359.69,99.67) .. (373.5,99.67) .. controls (387.31,99.67) and (398.5,110.86) .. (398.5,124.67) .. controls (398.5,138.47) and (387.31,149.67) .. (373.5,149.67) .. controls (359.69,149.67) and (348.5,138.47) .. (348.5,124.67) -- cycle ;
\draw  [color={rgb, 255:red, 155; green, 155; blue, 155 }  ,draw opacity=1 ][fill={rgb, 255:red, 128; green, 128; blue, 128 }  ,fill opacity=0.37 ] (432.8,125.6) .. controls (432.8,111.79) and (443.99,100.6) .. (457.8,100.6) .. controls (471.61,100.6) and (482.8,111.79) .. (482.8,125.6) .. controls (482.8,139.41) and (471.61,150.6) .. (457.8,150.6) .. controls (443.99,150.6) and (432.8,139.41) .. (432.8,125.6) -- cycle ;
\draw  [color={rgb, 255:red, 155; green, 155; blue, 155 }  ,draw opacity=1 ][fill={rgb, 255:red, 128; green, 128; blue, 128 }  ,fill opacity=0.37 ] (151.8,125.6) .. controls (151.8,111.79) and (162.99,100.6) .. (176.8,100.6) .. controls (190.61,100.6) and (201.8,111.79) .. (201.8,125.6) .. controls (201.8,139.41) and (190.61,150.6) .. (176.8,150.6) .. controls (162.99,150.6) and (151.8,139.41) .. (151.8,125.6) -- cycle ;
\draw  [color={rgb, 255:red, 155; green, 155; blue, 155 }  ,draw opacity=1 ][fill={rgb, 255:red, 128; green, 128; blue, 128 }  ,fill opacity=0.37 ] (169.8,125.6) .. controls (169.8,111.79) and (180.99,100.6) .. (194.8,100.6) .. controls (208.61,100.6) and (219.8,111.79) .. (219.8,125.6) .. controls (219.8,139.41) and (208.61,150.6) .. (194.8,150.6) .. controls (180.99,150.6) and (169.8,139.41) .. (169.8,125.6) -- cycle ;
\draw    (109.8,125.6) -- (149.8,125.6) ;
\draw [shift={(151.8,125.6)}, rotate = 180] [fill={rgb, 255:red, 0; green, 0; blue, 0 }  ][line width=0.75]  [draw opacity=0] (8.93,-4.29) -- (0,0) -- (8.93,4.29) -- cycle    ;

\draw  [dash pattern={on 4.5pt off 4.5pt}]  (239.5,124.67) -- (257.8,124.61) ;
\draw [shift={(259.8,124.6)}, rotate = 539.81] [fill={rgb, 255:red, 0; green, 0; blue, 0 }  ][line width=0.75]  [draw opacity=0] (8.93,-4.29) -- (0,0) -- (8.93,4.29) -- cycle    ;

\draw    (310.8,124.6) -- (346.5,124.66) ;
\draw [shift={(348.5,124.67)}, rotate = 180.1] [fill={rgb, 255:red, 0; green, 0; blue, 0 }  ][line width=0.75]  [draw opacity=0] (8.93,-4.29) -- (0,0) -- (8.93,4.29) -- cycle    ;

\draw    (398.5,124.67) -- (430.8,124.6) ;
\draw [shift={(432.8,124.6)}, rotate = 539.89] [fill={rgb, 255:red, 0; green, 0; blue, 0 }  ][line width=0.75]  [draw opacity=0] (8.93,-4.29) -- (0,0) -- (8.93,4.29) -- cycle    ;

\draw  [dash pattern={on 0.75pt off 0.75pt}]  (124.5,126.67) .. controls (126.49,128.54) and (126.87,130.68) .. (125.64,133.09) .. controls (124.24,134.5) and (124.6,136.04) .. (126.73,137.71) .. controls (128.61,138.26) and (129.04,139.74) .. (128.02,142.16) .. controls (127.09,144.64) and (127.59,146.07) .. (129.5,146.45) .. controls (131.77,147.66) and (132.52,149.49) .. (131.76,151.92) .. controls (130.66,153.51) and (131.29,154.82) .. (133.66,155.83) .. controls (136.02,156.69) and (136.95,158.35) .. (136.45,160.8) .. controls (135.54,162.49) and (136.3,163.66) .. (138.73,164.33) .. controls (141.14,164.86) and (142.24,166.35) .. (142.01,168.8) .. controls (141.3,170.57) and (142.18,171.62) .. (144.65,171.96) .. controls (147.08,172.17) and (148.32,173.5) .. (148.37,175.94) .. controls (148.54,178.41) and (149.86,179.65) .. (152.33,179.64) .. controls (154.09,178.95) and (155.13,179.82) .. (155.45,182.23) .. controls (155.88,184.66) and (157.33,185.74) .. (159.78,185.45) .. controls (162.18,185.06) and (163.69,186.04) .. (164.3,188.4) .. controls (165.03,190.76) and (166.2,191.44) .. (167.81,190.43) .. controls (170.16,189.79) and (171.77,190.61) .. (172.64,192.9) .. controls (173.61,195.18) and (175.26,195.92) .. (177.61,195.11) .. controls (179.9,194.22) and (181.17,194.72) .. (181.43,196.59) .. controls (182.62,198.78) and (184.36,199.36) .. (186.63,198.34) .. controls (188.84,197.25) and (190.61,197.74) .. (191.94,199.82) .. controls (193.37,201.87) and (194.72,202.19) .. (195.99,200.77) .. controls (198.12,199.5) and (199.94,199.84) .. (201.45,201.8) .. controls (203.06,203.73) and (204.91,203.99) .. (206.99,202.58) .. controls (208.09,201.03) and (209.49,201.16) .. (211.18,202.99) .. controls (212.95,204.79) and (214.82,204.9) .. (216.79,203.33) .. controls (217.78,201.7) and (219.19,201.73) .. (221.03,203.42) .. controls (222.94,205.07) and (224.83,205.04) .. (226.68,203.32) .. controls (228.47,201.56) and (229.88,201.48) .. (230.93,203.08) .. controls (232.97,204.59) and (234.85,204.41) .. (236.57,202.55) .. controls (238.22,200.66) and (239.63,200.48) .. (240.79,202) .. controls (242.94,203.34) and (244.81,203.02) .. (246.39,201.04) .. controls (247.89,199.03) and (249.28,198.74) .. (250.55,200.17) .. controls (252.79,201.35) and (254.62,200.89) .. (256.04,198.8) .. controls (257.38,196.69) and (258.74,196.29) .. (260.11,197.62) .. controls (262.43,198.63) and (264.21,198.03) .. (265.46,195.84) .. controls (266.62,193.63) and (268.37,192.96) .. (270.7,193.83) .. controls (272.21,195) and (273.5,194.45) .. (274.56,192.16) .. controls (275.53,189.87) and (277.2,189.07) .. (279.58,189.75) .. controls (282.01,190.36) and (283.24,189.7) .. (283.26,187.79) .. controls (284.01,185.44) and (285.6,184.5) .. (288.02,184.98) .. controls (290.48,185.38) and (292.01,184.37) .. (292.61,181.95) .. controls (292.38,180.07) and (293.49,179.26) .. (295.93,179.53) .. controls (298.41,179.71) and (299.83,178.57) .. (300.18,176.12) .. controls (300.43,173.69) and (301.78,172.47) .. (304.23,172.48) .. controls (306.04,173.05) and (307,172.09) .. (307.11,169.62) .. controls (307.1,167.18) and (308.31,165.84) .. (310.74,165.61) .. controls (313.19,165.26) and (314.32,163.85) .. (314.12,161.39) .. controls (313.31,159.68) and (314.09,158.58) .. (316.48,158.08) .. controls (318.88,157.46) and (319.85,155.93) .. (319.38,153.5) .. controls (318.81,151.12) and (319.67,149.52) .. (321.98,148.7) .. controls (324.29,147.75) and (324.87,146.5) .. (323.73,144.97) .. controls (322.9,142.66) and (323.58,140.94) .. (325.79,139.83) .. controls (327.98,138.57) and (328.55,136.79) .. (327.5,134.48) -- (327.88,133.11) -- (329.41,126.08) ;
\draw [shift={(329.65,124.63)}, rotate = 458.87] [fill={rgb, 255:red, 0; green, 0; blue, 0 }  ][line width=0.75]  [draw opacity=0] (8.93,-4.29) -- (0,0) -- (8.93,4.29) -- cycle    ;

\draw  [color={rgb, 255:red, 155; green, 155; blue, 155 }  ,draw opacity=1 ][fill={rgb, 255:red, 128; green, 128; blue, 128 }  ,fill opacity=0.37 ] (189.5,124.67) .. controls (189.5,110.86) and (200.69,99.67) .. (214.5,99.67) .. controls (228.31,99.67) and (239.5,110.86) .. (239.5,124.67) .. controls (239.5,138.47) and (228.31,149.67) .. (214.5,149.67) .. controls (200.69,149.67) and (189.5,138.47) .. (189.5,124.67) -- cycle ;

\draw (128,109) node [scale=1.44] [align=left] {A};
\draw (326,108) node [scale=1.44] [align=left] {B};
\draw (413,108) node [scale=1.44] [align=left] {C};
\draw (219,180) node [scale=1.44] [align=left] {\textit{{\large \textbf{enable}}}};

\end{tikzpicture}

\end{adjustbox}
    \caption{An \emph{opportunity chain}~\citep{hilton2005course}, where event $A$ \emph{enables} $B$ and $B$ causes $C$.}
    \label{fig:fig-opportunitychain}
\end{figure}


~\citeauthor{hilton2005course}~\citep{hilton2005course,mcclure2007judgments,hilton2010selecting} note that humans make use of \emph{opportunity chains} to describe events through causal explanation. An opportunity chain takes the form of $A$ enables $B$ and $B$ causes $C$ (depicted in Figure \ref{fig:fig-opportunitychain}), in which we call $B$ the `distal' event or action. For example, an accident can be caused by slipping on ice which was \emph{enabled} by water from a storm the day before. Opportunity chains are \emph{causal chains} that can be extracted from action influence models. Thus action influence models can be used as a platform to augment causal explanations with opportunity chains.

To that end, we propose a \emph{distal} explanation model that can generate opportunity chains as explanations for model-free RL agents. We provide definitions for distal explanations and learn the opportunity chains of extracted causal chains using a recurrent neural network~\citep{schuster1997bidirectional}. A distal explanations by itself would not make a \emph{complete} explanation. For this reason, we use action influence models~\citep{madumal2019explainable} to get the agent's `goals'. We further improve upon action influence models by using decision trees to represent the agent's policy.

\tm{Both here and in the experiments, I think we should use a language different to `three starcraft II scenarios. Both in AAAI and in IJCAI, at least one reviewer thought there were major limitations with the study because they just saw it as a game. But for the 2nd and 3rd scenarios, the participants weren't playing starcraft at all -- they were doing very different scenarios that just used starcraft as a simulator. Here, I would just say "three scenarios". Later in experiment design, we should make it clear that we used the simulation platform but created things that are not starcraft games}
\pmm{done}

We computationally evaluate the accuracy of task prediction~\cite[p.12]{hoffman2018metrics} and counterfactuals in 6 RL benchmark domains using 6 different RL algorithms, and show that our distal explanation model is robust and accurate across different environments and algorithms. Then we conduct human experiments using RL agents trained to solve 3 different scenarios, where agents solve 1) an adversarial task; 2) a search and rescue task; and 3) a human-AI collaborative build task. The human study was run with \textbf{90} participants, where we evaluate task prediction~\citep{hoffman2018metrics} and explanation satisfaction. Results indicate that our model performs better than the two tested baselines.

Our main contribution in this paper is twofold: 1) we introduce a distal action explanation model that is grounded on human data; 2) we extend action influence models by using decision trees to represent the agent's policy and formalise explanation generation from decision nodes and causal chains. As secondary contributions, we also provide the coded corpus of human-agent experiment with \textbf{240} explanations and two custom maps that are suited for explainability in the StarCraft II environment.

\section{Related Work}

In this section we discuss the body of literature that explores explainability in reinforcement learning agents. We also note work that have influenced explanation and interpretability of reinforcement learning, some of which are central to our own method. Further, we briefly overview human-centred literature on explanation which has greatly influenced this work.  

\subsection{Explanations of MDP based Agents}

Literature that sought to generate explanations for MDP based agents fall into the scope of preceding work on explainable RL. Often, these earlier work provided \emph{local} explanations in that the explanation is for a question about an action of the agent.

The concept of `relevant variables' in a factored state of an MDP was exploited by~\citeauthor{elizalde2007mdp}~(\citeyear{elizalde2007mdp}) to generate explanations. Explanations were primarily targeted at human trainees of a system and explanations were built-in and were presented when an the operator (trainee) selected an incorrect action. An explanation constitutes a relevant variable that is selected by an expert for each action.~\citeauthor{elizalde2008policy} later extended this work to generate explanations automatically based on the utility that a state variable had on the policy selecting the action~\citep{elizalde2008policy,elizalde2009expert}. ~\citeauthor{khan2009minimal}~(\citeyear{khan2009minimal}) was influenced from the relevant variable explanations and proposed minimally sufficient explanations for MDPs. Here, the long term effects of an optimal action is considered when generating the explanation.  Three domain independent templates were used as the basis of explanations. We later use one of these templates as a benchmark method in the evaluation section. Relevant variable explanations present a straightforward method of generating explanations from an MDP, though their inability provide contrastive explanations of counterfactuals remains a weakness.~\citeauthor{khan2009minimal}~(\citeyear{khan2009minimal}) attempted to remedy this by generating contrastive explanations through value-function comparisons. The effect MDP based agents' explanations have on `trust' was examined by \citeauthor{wang2016trust}~(\citeyear{wang2016trust}). Experiments were carried out to measure trust in human-robot teams influenced by Partially Observable MDP based explanations. As the measurement of trust was self-report, it is unclear whether the trust gain was from actually understanding the system.

\subsection{Policy Explanations}

Policy explanations make use of the agent's policy to extract explanations. Explanations can be at the local level or the global level. Global level explanations generally provide an explanation for the whole policy. Some studies suggest that humans are more receptive to global explanations of agents in certain situations~\citep{van2018contrastive}. We discuss literature on both global and local explanation methods in this section.

\tm{I'm not sure about the paragraph below: "the underlying model (i.e. an MDP model)". Is it the MDP model that is a black box, or the policy function? I understood that it is the policy function that is a black box}
\pmm{yes that is right, my mistake, fixed}

\citeauthor{struckmeier2019autonomous}~(\citeyear{struckmeier2019autonomous}) introduced a model-agnostic explanation generation method using agent policies. In cases where the underlying model (i.e. the policy function) is blackbox, \citeauthor{struckmeier2019autonomous} sample the policy of the agent to extract relevant state dimensions. Understanding of the agents' policies were measured in a human-experiment and the perceived understanding of the human participant was used as a proxy to show the transparency of the agent. 

Policy explanations of an agent generally aims to provide a `global' interpretation of the agent's behaviour. ~\citeauthor{hayes2017improving}~(\citeyear{hayes2017improving}) sought to improve the transparency by providing policy level explanations for agent based robot controllers. These behavioural explanations of the agent are considered as `summaries' of the agent's policy. Discreet, continuous and multi-agent domains were used to evaluate the generated policy descriptions against expert descriptions and were shown to improve the transparency of the robot.  ~\citeauthor{amir2018highlights}~(\citeyear{amir2018highlights}) also aims to summarise the agent's behaviour and introduced the HIGHLIGHTS algorithm. Important states are extracted from the agent's execution trace based on the Q-values. Human-subject experiments showed that participants preferred HIGHLIGHTS summary explanations compared to full policy explanations though in some situations participants' assessments did not always correlate with their confidence. Policy summarisation was also explored in the context of inverse reinforcement learning to investigate if these explanations are viable if there is a discrepancy between the agent's model and the human's mental model. 

\tm{above: " participants preferred HIGHLIGHTS summary explanations" -- compared to what? Should make this clear here}
\pmm{done}

\subsection{Reinforcement Learning Agent Explanations} 

Here we discuss work in recent years that specifically use characteristics of reinforcement learning (e.g. rewards) to explain behaviour. These characteristics can be used to create an approximate model of the agents policy or model and then generate explanations through using the approximate model.

\citeauthor{tabrez2019explanation}~(\citeyear{tabrez2019explanation}) proposed a framework (RARE) that repair the agent's understanding of the domain reward function through explanation. RARE is especially useful in human-agent collaborative scenarios when the human's reward function of the collaborative task is erroneous. Explanations are given to the collaborators to update their own reward function. Human experiments were conducted to demonstrate the effectiveness of the RARE framework in collaborative tasks. Explanation in the context of interactive reinforcement learning (IRL) has been studied~\citep{fukuchi2017autonomous}. This approach uses the instructions given in the IRL process to the agent as representations to generate explanations about the future behaviour of the agent. Evaluated through a human study, this method affirms that when explanations are given in a familiar medium to the human (e.g. using instruction representations) can yield a deeper understanding of the agent. ~\citeauthor{van2018contrastive}~(\citeyear{van2018contrastive}) developed a method that can translate an MDP of a RL agent to an interpretable MDP. This translation model can then be used to generate a \emph{contrastive} policy that can be queried using contrastive questions. A pilot study was carried out to evaluate the method, where the reported findings show that participants preferred the interpretable policy level explanations. Though these explanations were contrastive they were not based on an underlying causal model. Reward decomposition was used by ~\citeauthor{juozapaitis2019explainable}~(\citeyear{juozapaitis2019explainable}) to generate minimally sufficient explanations, where reward differences were used to provide explanations which answer what action does have an `advantage' over another. ~\citeauthor{juozapaitis2019explainable} utilise the nature of the reward structure often present in domains to explain action preferences of the agent.

\subsection{Decision Tree Policy Explanations}

Central to our own work, we discuss how interpretability and explainability was achieved through representing agents' policies as decision trees or graphs.

From early work that represented the agent policy as a decision tree using the `G' algorithm~\citep{chapman1991input}, past literature has explored how decision trees can be used to represent and abstract policies of MDPs. \citeauthor{roth2019conservative}~(\citeyear{roth2019conservative}) proposed a Q-improvment algorithm that builds an abstract decision tree policy for factored MDP based RL agents. Although decision tree policies are claimed to be more interpretable to humans than blackbox policies, the extent to which this is true is unclear as this work lacks human experiments. Abstract policy graphs have also been used as the basis to generate policy level explanations~\citep{topin2019generation}. A feature importance measure was used to abstract multiple states into an abstract state which is then used to build the policy graph. The interpetability of the graph was evaluated computationally that shows a linear growth of the explanation size against an exponential growth of state-space. Although this implicitly demonstrates the interpretabilty of the approach, human experiments are needed to understand the effectiveness of the method. 

Though above methods address interpretability to an extent, to the best of our knowledge previous literature has not studied how decision nodes from a decision tree can be incorporated \emph{with causal chains} to provide explanations that are human-centred.

\subsection{Human-Centred Explanation}

Some researchers have recently emphasised how humans models of explanations can benefit XAI systems~\citep{miller2018explanation} and how humans expect familiar models of explanations from XAI systems~\citep{de2017}. Though some recent progress has been made~\citep{madumal2019explainable}, human-centred computational models is still in its infancy.

~\citeauthor{hilton2005course}~\citep{hilton2005course,mcclure2007judgments,hilton2010selecting} has explored how causal chains of events inform and influence the explanations of humans. \emph{Opportunity chains} can inform the explainee about long term dependencies that events have on each other, where certain events \emph{enable} others. Human experiments have also been carried out that investigate he effects of opportunity chains on human-to-human explanation~\citep{hilton2005course}. However, this work has not yet been extended to the case of model-free MDPs.

Our proposed \emph{distal explanation} model take insights from social psychology literature to combine \emph{opportunity chains} with \emph{causal} explanations. To the best of our knowledge this is the first of such model in the context of explainable reinforcement learning agents.

\section{Human-Agent Study: Insights from Human Explanations}
\label{insights}

\tm{I would suggest making this section read a bit more like a study: outline the goals, then the setup, etc., all in their own section, rather than in a sub section. This gives the study a bit more of a proper feel. With the section being short and the study just being in Section 3.2, it feels a bit like a `throw-away' study, when in fact, it was a very good study that gave us some useful insight}
\pmm{done}

\liz{I suggest it also needs more explicit discussion as to the insight gained, that separates it from the Hilton et al (2010) study}
\pmm{added some more discussion  contrasting to Hilton 2010 }
\ls{I think the commentary re your use of distal (looking forward) versus that in the psych literature (looking backward) needs to be more visible; also, there seems to be a thread of work using the `distal' terminology that may warrant some visibility? eg see 2016 Cog Sci Society paper by Simon Stephan \url{https://scholar.google.de/citations?user=_GsFvvAAAAAJ} ; and maybe also you need a bit more on contemporary work on human models given the subsection title (which probably should be narrowed anyway to use the term "causal" somewhere? eg  \url{http://cicl.stanford.edu/publication/gerstenberg2020csm/} on counterfactual simulation and maybe a reference to Tania Lambrozo's work on causal explanation?}
\pmm{I have added description of looking backward and forward, also thanks for pointing ut Stephans work, I was not aware of it. have also cited Lambrozo's work. I think Gerstenberg's paper doesnt really fit well in this section as it's about counterfactual simulation and judgment. But I have changed the section name for causal explanations}

In this section, we discuss insights we can gain from human models of explanation in literature. We then ground these models in data by conducting a human-agent experiment.  

\subsection{Human Models of Causal Explanation}

\emph{Causality} is a recurring concept in explanation models of social psychology and cognitive science literature~\citep{hilton1990conversational,hornsby1993agency,lombrozo2017causal}. Using causal models as the basis for explanation seems natural and intuitive to humans~\citep{sloman2005causal}, since we build causal models to represent the world and to reason about it. Thus, it is plausible that, when used in intelligent agents, causal models have the ability to provide `good' explanations to humans. 

\tm{I suggest outlining the five different types of explanations that Hilton defines}
\pmm{done}

Importantly, causal models consist of \emph{causal chains}. A causal chain is a path that connects a set of events, where a path from event $A$ to event $B$ indicates that $A$ has to occur before $B$~\citep{miller2018explanation} (we use event and action interchangeably in the paper).~\citeauthor{hilton2005course} (~\citeyear{hilton2005course}) define five types of causal chains that lead to five different types of explanations. ~\citeauthor{hilton2005course} categorise these as, temporal, coincidental, unfolding, opportunity chains and pre-emptive. Through human experiments, ~\citeauthor{nagel2016explanations} (\citeyear{nagel2016explanations}) demonstrated that \emph{distal} causes forms significant portion of an explainee's understanding of a terminal cause. ~\citeauthor{bohm2015people} (\citeyear{bohm2015people}) also affirms that, humans give both proximal and distal causes as explanations. Its important to note that, while in cognitive psychology literature a distal cause is a remote cause of an event in the past (essentially looking `backward' from an event), in our agent simulations we use the distal terminology to denote an `action' that is remote in the future (according to the agent's viewpoint this is looking `forward' from a present event/action).~\citeauthor{hilton2010selecting}~(\citeyear{hilton2010selecting}) also explored how humans select different causal chains to provide explanations through human experiments. We conduct a similar study to gain insights from human models of explanation in a human-agent setting, and report results below.

\subsection{Study Objectives}

We seek to investigate how humans provide explanations of intelligent agents' behaviour and what concepts are present in such explanations. In contrast to similar studies done in social psychology~\citep{hilton2010selecting}, our experiments present explanations of the agent's behaviour first to the participant and then gives the freedom to form their own explanations of the agent. The main objective of the study is to discover the frequency of different concepts in these human generated explanations given the agent behaviour explanations using different explanation methods.

\begin{table}[!ht]
\centering
\caption{Codes (of the concepts) and descriptions of human generated explanations of agent behaviour. Examples are given from different participants.}
  \begin{tabular}{p{2cm}p{4cm}p{8cm}}
    \toprule
    Code & Description & Example\\
    \midrule
    Action & An \textbf{action} of the agent & P10: ``It will keep \textbf{attacking} while it has the advantage''  \\
    
    Feature & A \textbf{feature} of the agent & P4: ``The optimal number of \textbf{supply depots} is 2. and they should build those before a \textbf{barracks}'' \\
    
    Temporal & Refer some \textbf{temporal} quality & P12: ``I think the artificial player will want to train marines right away so that it has an army quickly and \textbf{be able to attack} the enemy.'' \\
    
    Objective & Refer to a short term \textbf{objective} & P12: ``I think the artificial player \textbf{will want to} train marines right away so that it has an army quickly and be able to attack the enemy.'' \\
    
    Causality & Implies a \textbf{causal} relationship & P3: ``\textbf{you need an army to attack}[action] and by training marines you can do that''  \\
    
    Quantitative & Refers to a \textbf{numerical} value of a \textbf{feature} & P4: ``The optimal number of supply depots is \textbf{2}. and they should build those before a barracks''  \\
    Qualitative & A \textbf{qualitative} reference to a \textbf{feature} & P4: ``\textbf{As long as they have enough} healthy marines. they should keep attacking'' \\
    
    Uncertainty & Mentions \textbf{uncertainty} & P5: ``The army is in good health. \textbf{it will most likely} continue to attack.'' \\
    
    Contrastive & \textbf{Contrasting} a \textbf{feature} with another & P15 ``There are 2 \textbf{supply depots but only 1 barrack.}'' \\
    Goal & Refer to the \textbf{goal(s)} of the agent & P10: ``The point of the game is \textbf{to kill the enemy} and \textbf{destroy their base}. so (incorrectly) the AI thinks the next step is to attack.'' \\
\bottomrule
  \end{tabular}

  \label{tablecodes}
\end{table}

\subsection{Experiment Design}
We conducted a human-agent study with 30 participants. In the first phase, participants were shown reinforcement learning agents playing the game StarCraft II. The agent behaviour (policy) was explained by providing `local' explanations of agents' actions using one of 3 different explanations models: 1) No explanations, just visual description of the agent's behaviour; 2) State-action based explanations~\citep{khan2009minimal}; and 3) Causal explanations~\citep{madumal2019explainable}. Participants were divided evenly for each of these explanation models. Experiment was run on a web based interactive interface in through the \emph{Amazon Mechanical Turk}~\citep{buhrmester2011amazon}.

In the second phase, participants were shown new agent behaviour and were asked to `predict' the agent's next action. Participants are expected to predict the next action based on the learned model of the agent in the first phase through explanations. This prediction task is not important to the objectives of this study, but is used as a way to get the participants to reasong about behaviour. In the same page, participants were then asked formulate their own explanations about the agent. Participants were given a text-box to input the formulated explanations with no restrictions to word limit. This process is repeated for 8 rounds.

\tm{How many participants were filtered? What was the threshold}
\pmm{mentioned}

To filter out devious participants, we used the following approaches. Explanations containing less than three words or gibberish text were omitted. We also considered the time it took to input the explanation as a threshold. We omitted six participants according to the above criteria. In total we obtained a total of 240 explanations.

\tm{I would say that we could have a bit more detail here. For example, note that the `prediction' part is not important to this study, etc. Discuss a bit more what they participants were asked to do and how explanations were filtered etc}
\pmm{done}

\tm{I'd suggest adding a table with the codes and their descriptions. Perhaps even just add a new column with the frequencies and remove the graph? Or you can keep the graph if you think it is useful, but I think a description of the codes is important}
\pmm{done}

\subsection{Method}

\begin{figure}[!t]
\centering
\includegraphics[width=\columnwidth]{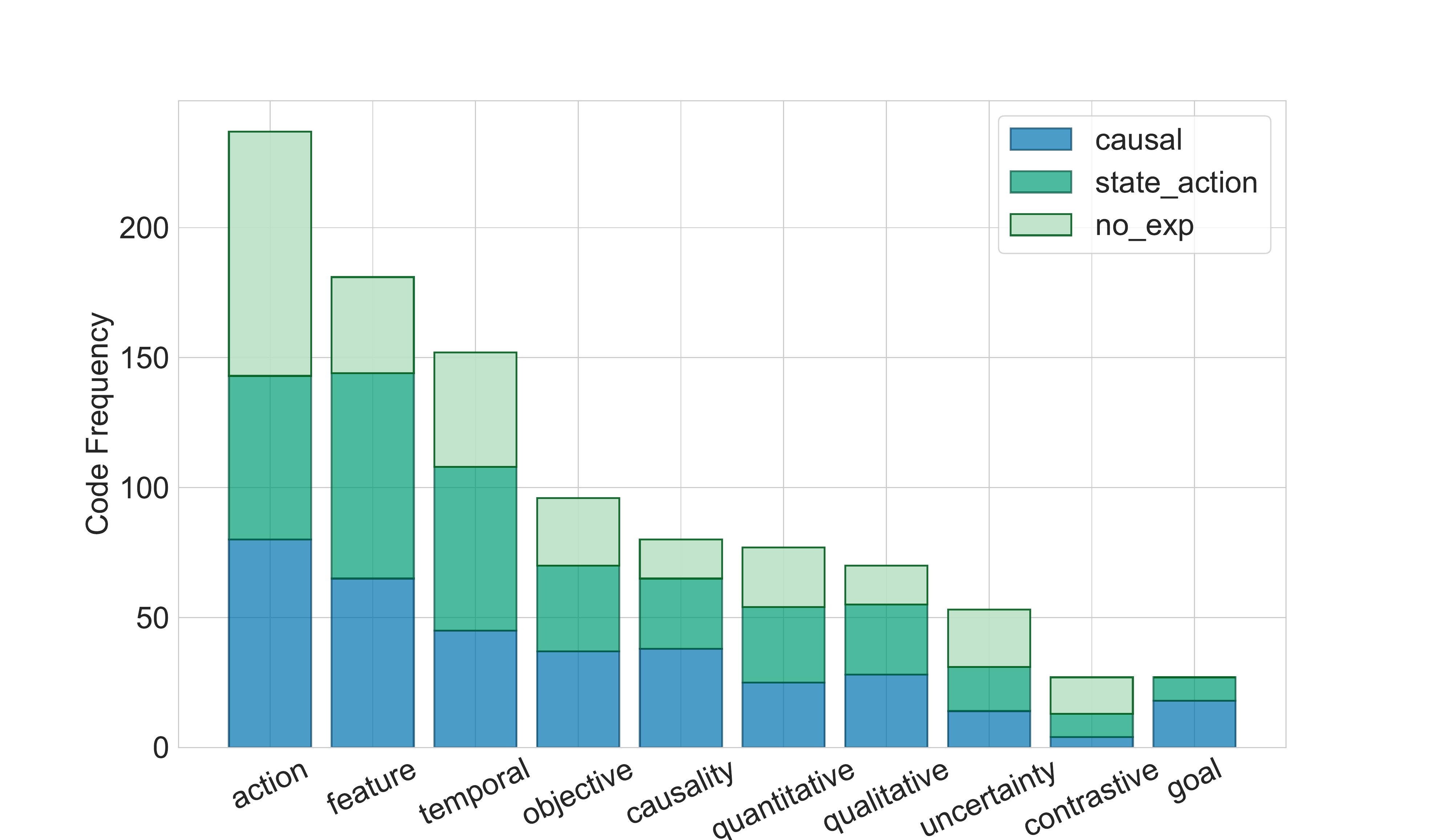}
\caption{Codes and their frequencies of 240 human explanations of reinforcement learning agents (that were using 3 different explanation models)} \label{fig1}
\end{figure}

\tm{As noted earlier, I think we should just say 'Thematic analysis', which is effectively one iteration of grounded theory}
\pmm{changed}

We use thematic analysis~\citep{braun2006using} to \emph{code} the data and to identify concepts. By using thematic analysis, meaningful insights can be gained on how explanations of agents behaviour relates to existing literature on human explanations. As the first step in the thematic analysis, each explanation will be be divided into small chunks to identify categories and then these will be divided further into codes. Intuitively, a `code' represents an atomic concept that exist in the explanation corpus. For an example, when a reference to an `action' of the agent is present in the explanation, the sub-string of that reference can be \emph{coded} (tagged) as an \textbf{Action}. This process is done manually until all the data chunks and explanations are coded. To ensure correctness, further passes through the explanation corpus is done as an attempt to identify new concepts that might have been missed in the first pass. Coded concepts and their descriptions are given in Table \ref{tablecodes}, along with example explanations extracted from participants.   


\subsection{Results}

\tm{It is not clear to me that discussing the three different explanation models is adding anything. What no just say that we conducted a study with 30 participants where they saw behavour etc. ,etc., and then had to do predictions and explain the predictions. The graph in FIgure 2 can then just add the total number across all three. While the conditions played an important part in the previous study, it is not so relevant for this particular purpose, and I think it just distracts the reader now, who may ask for more details on the three conditions, etc. There is not a major different between conditions to the point where it should influence our results}

\liz{I think it's risky to completely ignore that there were three explanation types -- but agree they should show on a single (stacked) bar chart - and suggest you order by decreasing height to make the description clearer}
\pmm{I have represented the bar chart as a stacked chart and added some more description in teh results section}

Figure \ref{fig1} shows the frequencies of 9 codes across the 3 explanation models of the RL agents. Participants referred to `actions' and `features' of the agent the most, and often included the `objective' or the `goal' of the agent, which is present in action influence models~\citep{madumal2019explainable}. Most importantly, the third most frequent code is `temporal', in which participants refer to future actions the agent will take (i.e. distal actions). For example, consider an explanation from the data corpus, ``The AI will want to have barracks so that it can then train soldiers to engage in attacks. It will want to progress''. Here, the participant's explanation contains the distal action `train soldiers' which is \emph{enabled} by `have barracks'. `Causality' is also present in the explanations, interestingly even in `No explanation' and State-action based explanation models. This suggests that humans frequently associate causal relationships when generating explanations. Our human-agent experimental data reaffirm the presence of opportunity chains in causal chains~\citep{hilton2005course}, and show that these are frequently used to express how future actions are dependent on current actions of agents. 

\tm{I would suggest that here or perhaps in the discussion, look at your previous work on action influence models and also other XRL work, and identify which methods refer to which things. Perhaps a table that lists the codes as columns and other XRL methods as rows, with a tick in the cell if such the concept in the code is used. I imagine we would see a column of causality mostly empty and a column of temporal entirely empty.}
\pmm{excellent suggestion about the table, I have added that}

\subsection{Discussion}

\begin{table}[!ht]
\centering

\caption{Presence of the concepts that were derived from codes, in different explainable reinforcement learning methods.}
\resizebox{\textwidth}{!}{%
  \begin{tabular}{p{2.5cm}cccccccccc}
    \toprule
    XRL Method & Action & Feature & Temporal & Causal & Contrast & Objec & Goal & Quan & Qual & Uncer  \\
    \midrule
    ~\cite{elizalde2007mdp} 	& \checkmark & \checkmark & ~ & ~ & ~ & ~ & ~ &  \checkmark & ~ & ~\\
    ~\cite{khan2009minimal} 	& \checkmark & \checkmark & ~ & ~ & \checkmark & ~ & ~ &  \checkmark & ~ & \checkmark\\
    ~\cite{wang2016trust} 	& ~ & \checkmark & ~ & ~ & ~ & \checkmark & ~ &  \checkmark & \checkmark & \checkmark\\
    ~\cite{van2018contrastive} 	& \checkmark & \checkmark & ~ & ~ & \checkmark & ~ & ~ &  \checkmark & ~ & ~\\
    ~\cite{struckmeier2019autonomous} 	& \checkmark & \checkmark & ~ & ~ & ~ & ~ & ~ &  \checkmark & ~ & ~\\
    ~\cite{hayes2017improving} 	& \checkmark & \checkmark & ~ & ~ & \checkmark & \checkmark & ~ &  \checkmark & ~ & ~\\
    ~\cite{amir2018highlights} 	& \checkmark & \checkmark & ~ & ~ & ~ & \checkmark & ~ &  \checkmark & ~ & ~\\
    ~\cite{tabrez2019explanation} 	& ~ & \checkmark & ~ & ~ & ~ & \checkmark & \checkmark &  \checkmark & ~ & ~\\
    ~\cite{fukuchi2017autonomous} 	& ~ & \checkmark & ~ & ~ & ~ & ~ & ~ &  \checkmark & ~ & ~\\
    ~\cite{juozapaitis2019explainable} 	& ~ & \checkmark & ~ & ~ & ~ & \checkmark & \checkmark &  \checkmark & ~ & ~\\
    ~\cite{madumal2019explainable} 	& \checkmark & \checkmark & ~ & \checkmark & \checkmark & \checkmark & \checkmark &  \checkmark & \checkmark & ~\\
    Proposed method 	& \checkmark & \checkmark & \checkmark & \checkmark & \checkmark & \checkmark & \checkmark &  \checkmark & \checkmark & ~\\
    
\bottomrule
  \end{tabular}%
}

  \label{tablecodecomparrison}
\end{table}


The concepts that were derived from the codes are present in previous explainable reinforcement learning methods to varying degrees. Table \ref{tablecodecomparrison} shows how these concepts are distributed. As most of these methods were not developed in a ground-up manner, some important concepts present in human explanations were not implemented in their explanation generation. When developing novel explanation models, insights gained from human-agent studies can help ground the model in the characteristics of \emph{human explanation}. Human grounded explainable models can be more effective and accepted when deployed~\citep{miller2018explanation,langley2017explainable}. To this end, we conducted human-agent experiments to discover how a human would explain the reasoning and behaviour of an agent, when the agent has given prior explanations of it's own actions. When these human explanations were abstracted into `codes', notable concepts like `causality' and `temporality' emerged. Previous work done in social psychology support our findings and coincide well with notions like opportunity chains~\citep{hilton2005course}.

Though previous studies have explored the structure of causal chains in human explanations~\citep{hilton2010selecting}, these are largely done in the absence of an intelligent agent. Further, in ~\citep{hilton2010selecting}, an explanation structure is investigated for events that have already occurred. In our study, as human explanations are for the \emph{behaviour} of the agent, they can refer to how the past and present actions of the agent can influence the future. Ultimately, we use the resultant concepts of \emph{causality} and \emph{distal opportunity chains} to propose the distal explanation model for reinforcement learning agents.

\section{Preliminaries}


In this section, we present the necessary background that is required to follow the remainder of the paper.

\subsection{Markov Decision Processes}
We concern ourselves with providing an explanations for Markov Decision Process (MDP) based model-free RL agents. An MDP is a tuple $\left ( \mathcal{S}, \mathcal{A}, \mathcal{T}, \mathcal{R},  \gamma   \right )$, where $\mathcal{S}$ and $\mathcal{A}$ give state and action spaces respectively (here we assume the state and action space is finite and state features are described by a set of variables $\phi$); $\mathcal{T} = \left \{ P_{sa} \right \}$ gives a set of state transition functions where $P_{sa}$ denotes state transition distribution of taking action $a$ in state $s$; $\mathcal{R} : \mathcal{S} \times \mathcal{A} \rightarrow \mathbb{R}$ is a reward function and $\gamma = \left [0, 1  \right )$ gives a discount factor. The objective of a reinforcement learning agent is to find a policy $\pi$ that maps states to actions maximizing the expected discounted sum of rewards. In model-free reinforcement learning, $\mathcal{T}$ and $\mathcal{R}$ is not known and the agent does not explicitly learn them.

\tm{I suggest defining mdoel-free RL here, as this is what we are tackling}
\pmm{done}

\subsection{Structural Causal Models}


Structural causal models (SCMs) \citep{halpern2005causes} provide a formalism for representing variables and \emph{causal} relationships between those variables. SCMs represent the world using random variables, divided into exogenous (external) and endogenous (internal), some of which might have causal relationships which each other. These relationships can be described with a set of \emph{structural equations}.


\begin{definition}\label{def0}
A \emph{signature} $\mathcal{S}$ is a tuple $\left ( \mathcal{U}, \mathcal{V}, \mathcal{R} \right )$, where $\mathcal{U}$ is the set of exogenous variables, $\mathcal{V}$ the set of endogenous variables, and $\mathcal{R}$ is a function that denotes the range of values for every variable $\mathcal{Y} \in   \mathcal{U} \cup \mathcal{V} $.\hfill\BlackBox
\end{definition}

\begin{definition}\label{def1}
A \emph{structural causal model} is a tuple $M = \left (\mathcal{S}, \mathcal{F}  \right )$, where $\mathcal{S}$ is as in Definition \ref{def0} and $\mathcal{F}$ denotes a set of structural equations, one for each $X \in \mathcal{V}$, such that $F_X : \left ( \times_{U \in \mathcal{U}}\mathcal{R}(U) \right )\times\left ( \times_{Y \in \mathcal{V} -\left \{ X \right \}}\mathcal{R}(Y) \right )\rightarrow \mathcal{R}(X)$ give the value of $X$ based on other variables in $\mathcal{U} \cup \mathcal{V}$. That is, the equation $F_X$ defines the value of $X$ based on some other variables in the model. \hfill\BlackBox
\end{definition}

\tm{I think adding a short example here would help illustrate both the concept of models, instantiations, and actual causes}
\pmm{added an example}

A \emph{context} $\vec{u}$ is a vector of unique values of each exogenous variable $u \in \mathcal{U}$. A \emph{situation} is defined as a model/context pair $\left (M, \vec{u}  \right )$. Given a situation $\left (M, \vec{u}  \right )$ an \emph{instantiation} of $M$ given $\vec{u}$ is defined by assigning all endogenous variables the values corresponding to those defined by their structural equations. 

An \emph{actual cause} of an event $\varphi$ is a vector of endogenous variables and their values such that there is some counterfactual context in which the variables in the cause are different and the event $\varphi$ does not occur. An explanation is those causes that an explainee does not already know. Following example gives perspective to the notions discussed above.

\begin{ex}
Consider the \emph{coffee task}~\citep{boutilier1995exploiting} where a robot has to deliver coffee to a user. The state consists of six binary variables, robot location ($L$), robot is wet ($W$), robot has umbrella ($Umb$), raining ($Rn$), robot has coffee ($C$) and user has coffee ($Usr$). Actions of the robot are \emph{go}, \emph{buy coffee}, \emph{get umbrella} and \emph{deliver coffee}. Then we can identify the set of \emph{endogenous} variables $\mathcal{U}$ as $L$, $W$, $Umb$, $C$ and $Usr$ because the values of these variables can be influenced by the actions of the robot. In contrast, variable the variable $Rn$ (raining) is an \emph{exogenous} ($\mathcal{V}$) variable, because it is not defined by a function. A \emph{signature} for this is generated by combining $\mathcal{U}$, $\mathcal{V}$ and the value range the variables can take (in this case either 0 or 1). Having the signature at hand, we can formulate a \emph{structural causal model} $M$ by identifying the set of functions $\mathcal{F}$ that describe causal relationships of state variables. Assuming there is only one such function, we can define it $F_{Usr} = C + L$. This implies that the variable `user has coffee' is causally influenced by variables `robot has coffee' and `robot location'. Model $M$ can be \emph{instantiated} by getting the current values of the state variables and applying them to the set of $\mathcal{F}$. The \emph{actual cause} of the event $Usr$ being true is the vector ($C = 1$, $L = 1$) as both of these variables needs to be true for the user to have the coffee.
\end{ex}

For a more complete review of SCM's we direct the reader to~\citep{halpern2005causes}.

\tm{We will also need to add interventions and in particular the syntax $M_{vec{X}\leftarrow \vec{x}}$, which is used later}
\pmm{not exactly sure what is missing from the definitions}

\subsection{Action Influence Models}

Action influence models~\citep{madumal2019explainable} provide explanations of the agent's behaviour based on the knowledge of how actions influence the environment. Informally, action influence models are an extension of SCMs that are augmented with agent actions. These models capture the causal relationships that exist in agent's knowledge about the world (i.e. state variables). Action influence models are formally defined for RL agents as follows,

\begin{definition}\label{signature_action_influence}
Formally, a signature $S_{a}$ for an action influence model is a tuple $(\mathcal{U}, \mathcal{V}, \mathcal{R}, \mathcal{A})$, in which $\mathcal{U}$, $\mathcal{V}$, and $\mathcal{R}$ are as in SCMs from Defintion \ref{def0}, and $\mathcal{A}$ is the set of actions from an MDP.
\hfill\BlackBox
\end{definition}

\begin{definition}\label{def2}
An \emph{action influence model} is a tuple $\left (  S_a, \mathcal{F}\right)$, where $S_a$ is as above, and $\mathcal{F}$ is the set of structural equations, in which we have multiple for each $X \in \mathcal{V}$ --- one for each \emph{unique} action set that influences $X$. A function $F_{X.A}$, for $A \in \mathcal{A}$, defines the causal effect on $X$ from applying action $A$. The set of \emph{reward variables} $R \subseteq \mathcal{V}$ are defined by the set of nodes with an out-degree of 0; that is, the set of sink nodes.
\hfill\BlackBox
\end{definition}

\begin{definition}\label{def:actual_instantiation}
The \emph{actual instantiation}~\cite{madumal2019explainable} of an action influence graph is defined as $M_{\vec{\mathcal{V}}\leftarrow \vec{S}}$, in which $\vec{S}$ is the vector of state variable values from an MDP and $\mathcal{V}$ as in Definition \ref{signature_action_influence}. A \emph{counterfactual instantiation} for a counterfactual action $B$ is a model $M_{\vec{Z}\leftarrow \vec{S_Z}}$, where $\vec{Z}$ gives the instantiation of a counterfactual state $\vec{S_Z}$.
\hfill\BlackBox
\end{definition}

\tm{I don't think the counterfactual $M_{\vec{Z}\leftarrow \vec{S_Z}}$ has been defined? This could be either defined here or in the SCM background}
\pmm{I think I defined it in the above defintion}
\tm{True, but I'm not sure I would understand what it means without already knowing it from SCMs?}

In an \emph{actual instantiation}, we set the values of all state variables in the model, effectively making the exogenous variables irrelevant. Similarly, a \emph{counterfactual instantiation} assign values to the model $M$ that could have realised under the action $B$. 


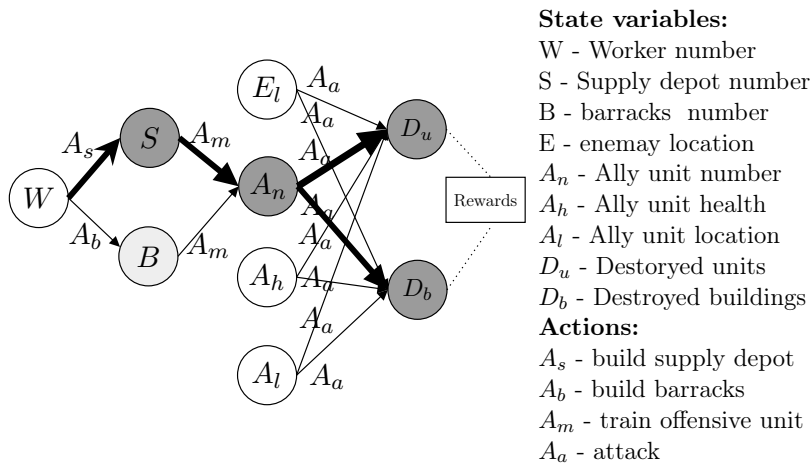
\begin{figure}[tb]

\centering
\resizebox {11cm} {!} {

\tikzset{every picture/.style={line width=0.75pt}} 

\begin{tikzpicture}[x=0.75pt,y=0.75pt,yscale=-1,xscale=1]

\draw  [color={rgb, 255:red, 0; green, 0; blue, 0 }  ,draw opacity=1 ] (3.8,120.6) .. controls (3.8,106.79) and (14.99,95.6) .. (28.8,95.6) .. controls (42.61,95.6) and (53.8,106.79) .. (53.8,120.6) .. controls (53.8,134.41) and (42.61,145.6) .. (28.8,145.6) .. controls (14.99,145.6) and (3.8,134.41) .. (3.8,120.6) -- cycle ;
\draw  [color={rgb, 255:red, 0; green, 0; blue, 0 }  ,draw opacity=1 ][fill={rgb, 255:red, 104; green, 104; blue, 104 }  ,fill opacity=0.11 ] (96.53,170) .. controls (96.53,156.19) and (107.73,145) .. (121.53,145) .. controls (135.34,145) and (146.53,156.19) .. (146.53,170) .. controls (146.53,183.81) and (135.34,195) .. (121.53,195) .. controls (107.73,195) and (96.53,183.81) .. (96.53,170) -- cycle ;
\draw  [color={rgb, 255:red, 0; green, 0; blue, 0 }  ,draw opacity=1 ][fill={rgb, 255:red, 155; green, 155; blue, 155 }  ,fill opacity=1 ] (198,110.67) .. controls (198,96.86) and (209.19,85.67) .. (223,85.67) .. controls (236.81,85.67) and (248,96.86) .. (248,110.67) .. controls (248,124.47) and (236.81,135.67) .. (223,135.67) .. controls (209.19,135.67) and (198,124.47) .. (198,110.67) -- cycle ;
\draw  [color={rgb, 255:red, 0; green, 0; blue, 0 }  ,draw opacity=1 ][fill={rgb, 255:red, 155; green, 155; blue, 155 }  ,fill opacity=1 ] (97.73,69.6) .. controls (97.73,55.79) and (108.93,44.6) .. (122.73,44.6) .. controls (136.54,44.6) and (147.73,55.79) .. (147.73,69.6) .. controls (147.73,83.41) and (136.54,94.6) .. (122.73,94.6) .. controls (108.93,94.6) and (97.73,83.41) .. (97.73,69.6) -- cycle ;
\draw  [color={rgb, 255:red, 0; green, 0; blue, 0 }  ,draw opacity=1 ] (197,187.67) .. controls (197,173.86) and (208.19,162.67) .. (222,162.67) .. controls (235.81,162.67) and (247,173.86) .. (247,187.67) .. controls (247,201.47) and (235.81,212.67) .. (222,212.67) .. controls (208.19,212.67) and (197,201.47) .. (197,187.67) -- cycle ;
\draw  [color={rgb, 255:red, 0; green, 0; blue, 0 }  ,draw opacity=1 ] (197,28.67) .. controls (197,14.86) and (208.19,3.67) .. (222,3.67) .. controls (235.81,3.67) and (247,14.86) .. (247,28.67) .. controls (247,42.47) and (235.81,53.67) .. (222,53.67) .. controls (208.19,53.67) and (197,42.47) .. (197,28.67) -- cycle ;
\draw  [color={rgb, 255:red, 0; green, 0; blue, 0 }  ,draw opacity=1 ] (197,270.67) .. controls (197,256.86) and (208.19,245.67) .. (222,245.67) .. controls (235.81,245.67) and (247,256.86) .. (247,270.67) .. controls (247,284.47) and (235.81,295.67) .. (222,295.67) .. controls (208.19,295.67) and (197,284.47) .. (197,270.67) -- cycle ;
\draw  [color={rgb, 255:red, 0; green, 0; blue, 0 }  ,draw opacity=1 ][fill={rgb, 255:red, 155; green, 155; blue, 155 }  ,fill opacity=1 ] (324.33,198.33) .. controls (324.33,184.53) and (335.53,173.33) .. (349.33,173.33) .. controls (363.14,173.33) and (374.33,184.53) .. (374.33,198.33) .. controls (374.33,212.14) and (363.14,223.33) .. (349.33,223.33) .. controls (335.53,223.33) and (324.33,212.14) .. (324.33,198.33) -- cycle ;
\draw  [color={rgb, 255:red, 0; green, 0; blue, 0 }  ,draw opacity=1 ][fill={rgb, 255:red, 155; green, 155; blue, 155 }  ,fill opacity=1 ] (323.67,62.33) .. controls (323.67,48.53) and (334.86,37.33) .. (348.67,37.33) .. controls (362.47,37.33) and (373.67,48.53) .. (373.67,62.33) .. controls (373.67,76.14) and (362.47,87.33) .. (348.67,87.33) .. controls (334.86,87.33) and (323.67,76.14) .. (323.67,62.33) -- cycle ;
\draw  [color={rgb, 255:red, 0; green, 0; blue, 0 }  ,draw opacity=1 ] (373.5,103) -- (440.5,103) -- (440.5,141.2) -- (373.5,141.2) -- cycle ;
\draw [color={rgb, 255:red, 0; green, 0; blue, 0 }  ,draw opacity=1 ][line width=0.75]    (53.8,120.6) -- (96.05,160.62) ;
\draw [shift={(97.5,162)}, rotate = 223.45] [fill={rgb, 255:red, 0; green, 0; blue, 0 }  ,fill opacity=1 ][line width=0.75]  [draw opacity=0] (8.93,-4.29) -- (0,0) -- (8.93,4.29) -- cycle    ;

\draw [color={rgb, 255:red, 0; green, 0; blue, 0 }  ,draw opacity=1 ][line width=3.75]    (53.8,120.6) -- (93.82,74.15) ;
\draw [shift={(97.73,69.6)}, rotate = 490.74] [fill={rgb, 255:red, 0; green, 0; blue, 0 }  ,fill opacity=1 ][line width=3.75]  [draw opacity=0] (22.33,-10.72) -- (0,0) -- (22.33,10.73) -- (14.83,0) -- cycle    ;

\draw [color={rgb, 255:red, 0; green, 0; blue, 0 }  ,draw opacity=1 ][line width=3.75]    (147.73,69.6) -- (193.35,106.87) ;
\draw [shift={(198,110.67)}, rotate = 219.25] [fill={rgb, 255:red, 0; green, 0; blue, 0 }  ,fill opacity=1 ][line width=3.75]  [draw opacity=0] (20.54,-9.87) -- (0,0) -- (20.54,9.87) -- cycle    ;

\draw [color={rgb, 255:red, 0; green, 0; blue, 0 }  ,draw opacity=1 ][line width=0.75]    (146.53,170) -- (196.69,112.18) ;
\draw [shift={(198,110.67)}, rotate = 490.94] [fill={rgb, 255:red, 0; green, 0; blue, 0 }  ,fill opacity=1 ][line width=0.75]  [draw opacity=0] (8.93,-4.29) -- (0,0) -- (8.93,4.29) -- cycle    ;

\draw [color={rgb, 255:red, 0; green, 0; blue, 0 }  ,draw opacity=1 ]   (247,187.67) -- (322.62,64.04) ;
\draw [shift={(323.67,62.33)}, rotate = 481.45] [fill={rgb, 255:red, 0; green, 0; blue, 0 }  ,fill opacity=1 ][line width=0.75]  [draw opacity=0] (8.93,-4.29) -- (0,0) -- (8.93,4.29) -- cycle    ;

\draw [color={rgb, 255:red, 0; green, 0; blue, 0 }  ,draw opacity=1 ]   (247,187.67) -- (322.35,198.06) ;
\draw [shift={(324.33,198.33)}, rotate = 187.85] [fill={rgb, 255:red, 0; green, 0; blue, 0 }  ,fill opacity=1 ][line width=0.75]  [draw opacity=0] (8.93,-4.29) -- (0,0) -- (8.93,4.29) -- cycle    ;

\draw [color={rgb, 255:red, 0; green, 0; blue, 0 }  ,draw opacity=1 ][line width=3.75]    (248,110.67) -- (320.39,193.81) ;
\draw [shift={(324.33,198.33)}, rotate = 228.95] [fill={rgb, 255:red, 0; green, 0; blue, 0 }  ,fill opacity=1 ][line width=3.75]  [draw opacity=0] (20.54,-9.87) -- (0,0) -- (20.54,9.87) -- cycle    ;

\draw [color={rgb, 255:red, 0; green, 0; blue, 0 }  ,draw opacity=1 ]   (247,28.67) -- (323.5,196.51) ;
\draw [shift={(324.33,198.33)}, rotate = 245.5] [fill={rgb, 255:red, 0; green, 0; blue, 0 }  ,fill opacity=1 ][line width=0.75]  [draw opacity=0] (8.93,-4.29) -- (0,0) -- (8.93,4.29) -- cycle    ;

\draw [color={rgb, 255:red, 0; green, 0; blue, 0 }  ,draw opacity=1 ]   (247,270.67) -- (322.87,199.7) ;
\draw [shift={(324.33,198.33)}, rotate = 496.91] [fill={rgb, 255:red, 0; green, 0; blue, 0 }  ,fill opacity=1 ][line width=0.75]  [draw opacity=0] (8.93,-4.29) -- (0,0) -- (8.93,4.29) -- cycle    ;

\draw [color={rgb, 255:red, 0; green, 0; blue, 0 }  ,draw opacity=1 ]   (247,270.67) -- (322.98,64.21) ;
\draw [shift={(323.67,62.33)}, rotate = 470.2] [fill={rgb, 255:red, 0; green, 0; blue, 0 }  ,fill opacity=1 ][line width=0.75]  [draw opacity=0] (8.93,-4.29) -- (0,0) -- (8.93,4.29) -- cycle    ;

\draw [color={rgb, 255:red, 0; green, 0; blue, 0 }  ,draw opacity=1 ]   (247,28.67) -- (321.84,61.53) ;
\draw [shift={(323.67,62.33)}, rotate = 203.71] [fill={rgb, 255:red, 0; green, 0; blue, 0 }  ,fill opacity=1 ][line width=0.75]  [draw opacity=0] (8.93,-4.29) -- (0,0) -- (8.93,4.29) -- cycle    ;

\draw [color={rgb, 255:red, 0; green, 0; blue, 0 }  ,draw opacity=1 ][line width=4.5]    (248,110.67) -- (317.77,66.1) ;
\draw [shift={(323.67,62.33)}, rotate = 507.43] [fill={rgb, 255:red, 0; green, 0; blue, 0 }  ,fill opacity=1 ][line width=4.5]  [draw opacity=0] (24.11,-11.58) -- (0,0) -- (24.11,11.58) -- cycle    ;

\draw [color={rgb, 255:red, 0; green, 0; blue, 0 }  ,draw opacity=1 ] [dash pattern={on 0.84pt off 2.51pt}]  (374.33,198.33) -- (411.5,145) ;

\draw [color={rgb, 255:red, 0; green, 0; blue, 0 }  ,draw opacity=1 ] [dash pattern={on 0.84pt off 2.51pt}]  (373.67,62.33) -- (411.5,103) ;

\draw (407,122.1) node [scale=1,color={rgb, 255:red, 0; green, 0; blue, 0 }  ,opacity=1 ] [align=left] {Rewards};
\draw (567,153) node [scale=1.44,color={rgb, 255:red, 0; green, 0; blue, 0 }  ,opacity=1 ] [align=left] {\textbf{State variables:}\\W - Worker number\\S - Supply depot number\\B - barracks \ number\\E - enemay location\\$A_n$ - Ally unit number\\$A_h$ - Ally unit health\\$A_l$ - Ally unit location\\$D_u$ - Destoryed units\\$D_b$ - Destroyed buildings\\\textbf{Actions:}\\$A_s$ - build supply depot\\$A_b$ - build barracks\\$A_m$ - train offensive unit\\$A_a$ - attack};
\draw (62,76) node [scale=1.7280000000000002,color={rgb, 255:red, 0; green, 0; blue, 0 }  ,opacity=1 ]  {$A_{s}$};
\draw (28.8,120.6) node [scale=1.7280000000000002,color={rgb, 255:red, 0; green, 0; blue, 0 }  ,opacity=1 ]  {$W$};
\draw (122.73,69.6) node [scale=1.7280000000000002,color={rgb, 255:red, 0; green, 0; blue, 0 }  ,opacity=1 ]  {$S$};
\draw (121.53,170) node [scale=1.7280000000000002,color={rgb, 255:red, 0; green, 0; blue, 0 }  ,opacity=1 ]  {$B$};
\draw (223,110.67) node [scale=1.7280000000000002,color={rgb, 255:red, 0; green, 0; blue, 0 }  ,opacity=1 ]  {$A_{n}$};
\draw (222,28.67) node [scale=1.7280000000000002,color={rgb, 255:red, 0; green, 0; blue, 0 }  ,opacity=1 ]  {$E_{l}$};
\draw (222,187.67) node [scale=1.7280000000000002,color={rgb, 255:red, 0; green, 0; blue, 0 }  ,opacity=1 ]  {$A_{h}$};
\draw (222,270.67) node [scale=1.7280000000000002,color={rgb, 255:red, 0; green, 0; blue, 0 }  ,opacity=1 ]  {$A_{l}$};
\draw (349.33,198.33) node [scale=1.44,color={rgb, 255:red, 0; green, 0; blue, 0 }  ,opacity=1 ]  {$D_{b}$};
\draw (348.67,62.33) node [scale=1.44,color={rgb, 255:red, 0; green, 0; blue, 0 }  ,opacity=1 ]  {$D_{u}$};
\draw (68,153) node [scale=1.7280000000000002,color={rgb, 255:red, 0; green, 0; blue, 0 }  ,opacity=1 ]  {$A_{b}$};
\draw (174,67) node [scale=1.7280000000000002,color={rgb, 255:red, 0; green, 0; blue, 0 }  ,opacity=1 ]  {$A_{m}$};
\draw (172,161) node [scale=1.7280000000000002,color={rgb, 255:red, 0; green, 0; blue, 0 }  ,opacity=1 ]  {$A_{m}$};
\draw (270,20) node [scale=1.7280000000000002,color={rgb, 255:red, 0; green, 0; blue, 0 }  ,opacity=1 ]  {$A_{a}$};
\draw (265,50) node [scale=1.7280000000000002,color={rgb, 255:red, 0; green, 0; blue, 0 }  ,opacity=1 ]  {$A_{a}$};
\draw (263,82) node [scale=1.7280000000000002,color={rgb, 255:red, 0; green, 0; blue, 0 }  ,opacity=1 ]  {$A_{a}$};
\draw (265,128) node [scale=1.7280000000000002,color={rgb, 255:red, 0; green, 0; blue, 0 }  ,opacity=1 ]  {$A_{a}$};
\draw (263,153) node [scale=1.7280000000000002,color={rgb, 255:red, 0; green, 0; blue, 0 }  ,opacity=1 ]  {$A_{a}$};
\draw (265,189) node [scale=1.7280000000000002,color={rgb, 255:red, 0; green, 0; blue, 0 }  ,opacity=1 ]  {$A_{a}$};
\draw (273,272) node [scale=1.7280000000000002,color={rgb, 255:red, 0; green, 0; blue, 0 }  ,opacity=1 ]  {$A_{a}$};
\draw (264,224) node [scale=1.7280000000000002,color={rgb, 255:red, 0; green, 0; blue, 0 }  ,opacity=1 ]  {$A_{a}$};

\end{tikzpicture}

}
\caption{Action influence graph of a StarCraft II agent~\citep{madumal2019explainable}} \label{action_influence_fig}
\end{figure}

Figure~\ref{action_influence_fig} shows the graphical representation of Definition \ref{def2} as an action influence graph of the StarCraft II agent described in the previous section, with exogenous variables hidden. These \emph{action influence models} are SCMs except that each edge is associated with an action. In the action influence model, each state variable has a \emph{set} of structural equations: one for each \emph{unique} incoming action. As an example, from Figure \ref{action_influence_fig}, variable $\hat{A_n}$ is causally influenced by $\hat{S}$ and $\hat{B}$ only when action $A_m$ is executed, thus the structural equation $\mathcal{F}_{A_n.A_m}\left ( S, B \right )$ captures that relationship.  

\tm{Each time I read this, I find the use of $A$ to indicate actions to be a bit confusing because it also represents ``Ally''. Perhaps use $\hat{S}$ or something instead of $A_s$ to distinguish actions from state variables?}
\pmm{done}





\subsection{Explanations}

An explanation is generally defined as a pair that contains; 1) an \emph{explanandum}, the event to be explained and 2) an \emph{explanan}, the subset of causes that explain that event~\citep{miller2018explanation}. In its simplest form, the explanation for the question `Why $P$?' would be in the form of `Because $Q$'. In the above example, $P$ is the explanandum and $Q$ is the explanan. As ~\citeauthor{lim2009and}~(\citeyear{lim2009and}) notes, \emph{why} and \emph{why not} questions are the most demanded explanatory questions. In the context of RL agents, we are interested in answering `Why $A$?' and `Why not $A$?' questions. Here, $A$ is an action of the agent and the explanation will be \emph{local}.

Action influence models can be used to generate \emph{minimally complete} explanations. An explanation that constitutes all the causes as an explanan risk overwhelming the explainee, thus it is important to balance the \emph{completeness} and the \emph{minimality} of the explanations ~\citep{miller2018explanation}. 

\tm{ON the presentation of explanations, I wonder if we should 'break rank' a bit from the previous paper, as it is always a cause for confusion. I can think of two things: (1) instead of $\left (  \vec{X_r}=\vec{x_r}, \vec{X_h}=\vec{x_h}, \vec{X_i}=\vec{x_i}\right )$, we could use $\left (  \vec{R}=\vec{r}, \vec{H}=\vec{h}, \vec{I}=\vec{i}\right )$, which just makes the text a bit clearer; and (2) There definitely needs to be some explanation of that these three sets of variables are, especially to focus on why the immediate predecessors of the reward nodes so important (it's not clear unless you understand the reward nodes are not variables). We also need to point back to the earlier figure. It is perhaps even worth pulling out the figure again and focusing on the chain specifically to make this definition very clear, but you may also be able to do this be just pointing back to it.}
\pmm{added an explanation and referred back to figure and changed the symbols}

\begin{definition}\label{ation_influence_minally_complete}
A \emph{minimally complete} explanation for an action $a$ under the actual instantiation $M_{\vec{\mathcal{V}}\leftarrow \vec{S}}$ is a tuple $\left (  \vec{R}=\vec{r}, \vec{H}=\vec{h}, \vec{I}=\vec{i}\right )$, in which $ \vec{R}$ is the vector of reward variables reached by following the causal chain of the graph to sink nodes; $\vec{H}$ the vector of variables of the head node of action $a$, $\vec{I}$ is the vector of variables that are immediate predecessors of any variable in $\vec{R}$ within the causal chain, with $\vec{r}$, $\vec{h}$, $\vec{i}$ giving the values of these variables under $M_{\vec{\mathcal{V}}\leftarrow \vec{S}}$ from Definition \ref{def:actual_instantiation}.
\hfill\BlackBox
\end{definition}

~\citet{mcclure1997you} argue that `goals' should be referred to in some form when explaining actions. In reinforcement learning, rewards of the agent can be thought of as a proxy for the goals. Though in most cases the `rewards' ($\vec{R}$ from Definition \ref{ation_influence_minally_complete}) on itself would not form a complete explanation, because they are not attached to variables. Immediate predecessor nodes ($\vec{I}$) of the reward nodes refer to the state variables that `trigger' rewards.. Though this combination now can explain the long term motivation of the agent, the head node ($\vec{H}$) attached to the action is used to explain the immediate (short-term) cause. From Figure \ref{action_influence_fig}, the explanation for \emph{Why action $A_s$} would constitute, $D_u$ and $D_b$ in as reward variables $\vec{R}$, $A_n$ in $\vec{I}$ and $S$ in $\vec{H}$.~\citet{madumal2019explainable} present a method for generating such explanations, and evaluate this on a large-scale user study.

\section{Distal Explanation Model}

From the insights gained from human explanations discussed in Section \ref{insights} we propose a distal explanation model that can generate explanations for opportunity chains. In the following sections we use the adversarial scenario (discussed at length in Section \ref{section5-1}) of the StarCraft II environment as a running example to aid the definitions.

\tm{I suggest breaking this into its own 'overview' section (Section 5.1), and spending a bit of time explaining the architecture: what each component does and how it fits together. Perhaps introduce a short illustrative example from the starcraft domain. At this point, the reader wouldn't really know why we are using a many-to-one sequence predictor. I would draw this in (using 'opportunity chains' terminiology too)}
\pmm{added the overview and expanded teh section explaining in more detail}

\subsection{Overview}

\begin{figure*}[!ht]
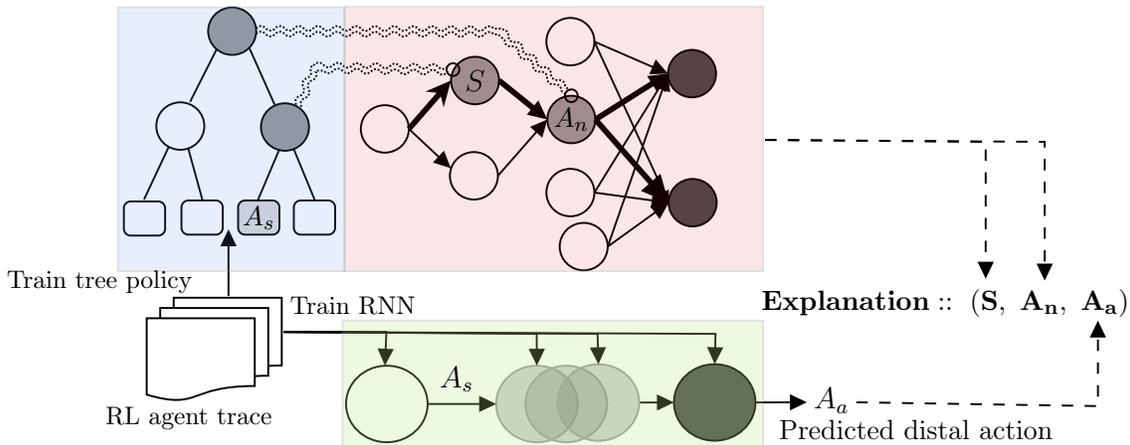

        \centering
        \resizebox{\textwidth}{!}{%

\tikzset{every picture/.style={line width=0.75pt}} 



        }
    \caption{An overview of the Distal explanation model}
    \label{fig:architecture}
    \end{figure*}

Figure \ref{fig:architecture} shows an overview of the distal explanation model. The model consists of four distinct components. First, state-action pairs are extracted as a replay dataset from the episodes during reinforcement learning. Dataset generation happens at the agent training time. This dataset is used to train the decision-tree policy (indicated as the blue sub-component in Figure \ref{fig:architecture}). The decision-tree policy is used as a surrogate policy for the agent, where it is used to extract reasons (in the form of decision nodes) for a given action (we discuss this process at length in Section 5.2). The dataset is also used to train the distal action predictor (shown as the green sub-component), which predicts dependent actions. Because we want to predict distal actions (contained in a opportunity chain) using a sequence of prior actions from the agent action trace, a many-to-one recurrent neural network~\citep{schuster1997bidirectional} is used as the predictor, though other sequence predictors can also be used. An action influence graph is used to extract causal chains (shown in red) that is used in conjunction with the decision tree policy to produce the final explanation.  The explanation is given as a three-tuple: reward nodes of the causal chain; matched decision nodes; and the predicted distal action.

Before formalising the distal explanation model we first discuss how explanations can be generated using decision tree policies.

\subsection{Causal Explanations from Decision Trees}


Although causal explanations from action influence models have been shown to perform better than state-action based~\citep{khan2009minimal} explanation models, the use of \emph{structural equations}~\citep{madumal2019explainable} models the environment rather than the policy of the agent. Thus, the explanations from these model why an action would be a good idea, rather than why the agent chose it. In this work, we instead propose to extract reasons for action selection from a \emph{surrogate} policy. We learn an interpretable surrogate policy in the form of a decision tree using batched replay data. If the agent's underlying policy is also a decision tree, this step can be omitted.

\tm{THe ref to Minh for experience replay is a bit strange as experience replay is much older than 2015}
\pmm{added Lin et al 1992 reference}

\begin{figure*}[!t]
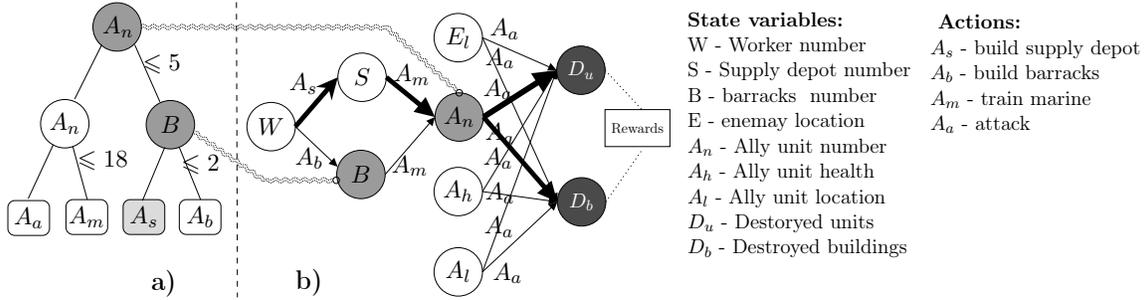

        \centering
        \resizebox{\textwidth}{!}{%


        }
    \caption{Generating explanations by mapping (a) decision nodes to (b) causal chains.}
    \label{fig:causal_nodes}
    \end{figure*}

\paragraph{Training The Surrogate Policy:} The \emph{distal explanation} model we introduce uses decision nodes of a decision tree that represent a surrogate policy to generate explanations with the aid of causal chains from an action influence model. Let $\mathbb{\widehat{T}}$ be a decision tree model. In each episode at the training of the RL agent, we perform experience replay~\citep{lin1992self} by saving $e_t = (s_t, a_t)$ at each time step $t$ in a data set $D_t = \left \{ e_1,..., e_t \right \}$. Drawing uniformly from $D$ as mini-batches, we train $\mathbb{\widehat{T}}$ using input $x = \vec{s}$ and output $y = \vec{a}$. Clearly, explanations generated from an unconstrained decision tree can overwhelm the explainee, as these produce a large number of decision nodes for a question. Thus we limit the growth of $\mathbb{\widehat{T}}$ by setting the max number of leaves to the number of actions in the domain (i.e. the leaves of the trained $\mathbb{\widehat{T}}$ will be the set of actions of the agent). We later show that this hardly affects the task prediction accuracy compared to a depth unconstrained decision tree for our experiments. To get the decision nodes of $\mathbb{\widehat{T}}$ in state $S_t$, we simply traverse the tree from the root node until we reach a leaf node and get the nodes of the path. The decision tree of the StarCraft II adversarial task is given in Figure \ref{fig:causal_nodes} a), with the decision nodes $A_n$ and $B$ for the action $A_s$. Each decision node maps to a feature variable of the agent's state. Figure \ref{fig:causal_nodes} shows how the decision nodes are mapped to the action influence graph, in the StarCraft II adversarial scenario.

\paragraph{Generating Explanations Using the Surrogate Policy:} In the context of an RL agent, we introduce a new definition of \emph{minimally complete} explanations using decision nodes for `why' questions below.

\tm{I think we are jumping forward a bit here. Perhaps first give the example of the 'explanation' from the decision tree only. Then explain that we filter out variables that are not relevant to the decision}
\pmm{added a description below}

A primitive explanation can be generated by using the decision-tree policy alone, by extracting the decision nodes of an action. E.g. for the question \emph{Why $A_s$}, we can obtain the decision nodes simply by traversing to the leaf node $A_s$ from the root node ($A_n$ and $B$ are the decision nodes in this case, as highlighted in Figure \ref{fig:causal_nodes} a)). However, an explanation like this can contain variables that are not \emph{causally} relevant to the action performed. This primitive explanation can be enhanced by taking the causal chain for the action being explained from an action influence model and filtering out causally irrelevant variables. We define this as a minimally complete explanation below.

\begin{definition}
\label{def_why_dt}
Given the set of decision nodes $\vec{X_d}=\vec{x_d}$ for the action $a$ from a decision tree $\mathbb{\widehat{T}}$, we define a \emph{minimally complete} explanation for a \emph{why} question as a pair $\left (  \vec{R}=\vec{r}, \vec{N}=\vec{n}\right )$, in which $ \vec{R}$ is the vector of reward variables reached by following the causal chain of the graph to sink nodes; $\vec{N}$ is such that $\vec{N}$ is the maximal set of variables in which $\vec{N} = (\vec{X_a}=\vec{x_a}) \cap (\vec{X_d}=\vec{x_d}) $, where $\vec{X_a}$ is the set of intermediate nodes of the causal chain of action $a$, with $\vec{r}$, $\vec{x_a}$ and $\vec{x_d}$ giving the values under the actual instantiation $M_{\vec{\mathcal{V}}\leftarrow \vec{S}}$ from Definition \ref{def:actual_instantiation}.
\hfill\BlackBox
\end{definition}

\tm{As with the earlier one, give a brief summary of what this definition means and why we do it; in particular, what the intersection is doing}
\pmm{explained a bit below}

Above definition only select the decision nodes (from the total set of decision nodes given from the decision-tree policy) that exist as intermediate nodes of the causal chain of the given action.

In the StarCraft II scenario, for the question `Why action $A_s$?', we can generate the minimally complete explanation by first finding the decision nodes for action $A_s$, shown as medium grey nodes in Figure \ref{fig:causal_nodes}(a). Then 
finding the causal chain of action $A_s$ (given by the bold path in Figure \ref{fig:causal_nodes}). And finally getting the common set of nodes from the causal chain and the decision nodes ($B$ in Figure \ref{fig:causal_nodes}) and appending the reward nodes ($D_u$ and $D_b$). Example \ref{compre_example_with_action_influence} below compare and contrast an explanation with and without the use of action influence models.

\tm{Give an example explanation? Perhaps even give a comparison to the explanation with and without the action influence graph?}
\pmm{given below}

\begin{ex}
\label{compre_example_with_action_influence}
Question: Why $A_s$?

\begin{center}
\noindent
\begin{tabular}{@{}l@{~~}p{0.6\textwidth}@{}}

\emph{Just decision-tree policy: } & Because Ally unit number ($A_n$) is 4 and Barracks number ($B$) is 1.\\

\emph{With action influence models: } & Because ally unit number ($A_n$) is 4 and the goal is to have more Destroyed Units ($D_u$) and Destroyed buildings ($D_b$).
\end{tabular}
\end{center}

\end{ex}

\subsection{Contrastive Explanations from Counterfactuals}

Counterfactuals explain events that did not happen---but could have under different circumstances. Counterfactuals are used to describe events from a `possible world' and to contrast them with what happened in actuality. Embedding these counterfactuals in explanations can make the explanation more meaningful~\citep{byrne2019counterfactuals}. Naturally, an explanation given to a `why not' question should compare the counterfactuals with the actual facts to form a \emph{contrastive explanation}~\citep{miller2018explanation, miller2018contrastive}. For this reason, we concern ourselves with generating contrastive explanations from decision nodes and causal models.

We generate the counterfactual decision nodes using Algorithm \ref{alg:contrastive}, in which we find the decision nodes of the counterfactual action $b$ by changing the decision boundary of the actual action $b$ in the decision tree. We can now define \emph{minimally complete contrastive} explanations for `why not' questions using these counterfactual decision nodes. 

\tm{First, I'm not entirely convinced we need this algorithm. Couldn't we just take the counterfactual action and get the path(s) to the root of the decision tree from that action, then select e.g. the one with the smallest number of changes nodes? Second, the algorithm does not seem to find the counterfactual unless the delta is large enough. Perhaps `moveBoundary' does this, but it does not consider that we may need to move more than one variable to get the counterfactual. In short, it's not clear what moveBOundary does, and if it just moves by a Delta, what if the delta change is not enough to get the counterfactual? It doesn't seem to do it again? Third, if you DO use this counterfactual approach, is there a reason the surrogate needs to be a decision tree? Couldn't we just use any counterfactual generation algorithm for a black-box model?}
\pmm{For the first question: when we have the counterfactual action, yes we can get the counterfactual decision nodes, but we cant get the counterfactual state, so we can't get the values for those decision nodes(values would be based on the current state), so we need to change the decision boundary to see the counterfactual values for the counterfactual decision nodes. Second question: delta should be adjusted accordingly to the domain I think, you can set it to a very small value, it just means the algorithm will be very slow, moveboundary change the state value by a delta value until the decision boundary changes and assign that final value as $x_m$, essentially there is a loop inside with a condition. For the third question: we can use other counterfactual generation algorithms yes to get the set counterfactual features that matters, doesnt have to be a decision tree, or decision nodes as we are simply comparing that together with nodes in the causal chain of action influence models}

\begin{algorithm}[!tb]
\small
\caption{Generating Counterfactuals}
\label{alg:contrastive}
\textbf{Input}: causal model $\mathcal{M}$, current state $S_t$, trained decision tree $\mathbb{\widehat{T}}$, \emph{actual} action $a$, $\Delta$\\
\textbf{Output}: contrastive explanation t $\vec{X_c}$
\begin{algorithmic}[1] 
\State $\vec{X_d} \leftarrow \mathbb{\widehat{T}} \cdot traversetree(a) $ ;vector of decision nodes of $a$ from $\mathbb{\widehat{T}}$
\State $\vec{X_c} \leftarrow [] $ ;vector of counterfactual decision nodes.

\For{every $D \in \vec{X_d}$ }
\State ${x_d} \leftarrow D \cdot decisionNodeValue() $; decision boundary value of $D$

\State ${x_{m}} \leftarrow D \cdot moveBoundary(x_d) $; boundary value changed by a $\Delta$.

\State ${S_tm} \leftarrow {S_t} \cup {x_{m}} $; modify the corresponding state feature variables with the new $x_{m}$.

\State $\vec{X_c} \leftarrow \vec{X_c} \cup \mathbb{\widehat{T}} \cdot predict({S_tm}) $; get the counterfactual decision nodes by getting the counterfactual action and then traversing the tree.

\EndFor

\State \Return{$\vec{X_c}$}

\end{algorithmic}
\end{algorithm}

\begin{definition}
\label{def_whynot_dt}
Given the set of decision nodes $\vec{X_d}=\vec{x_d}$ for the action $a$ from a decision tree $\mathbb{\widehat{T}}$, a \emph{minimally complete contrastive} explanation for a \emph{why not} question is a pair $\left (  \vec{R}=\vec{r}, \vec{X_{con}}=\vec{x_{con}}\right )$, in which $ \vec{R}$ is same as in Definition \ref{def_why_dt}; $\vec{X_{con}}$ is such that $\vec{X_{con}}$ is the maximal set of variables in which $\vec{X_{con}} = (\vec{X_b}=\vec{x_b}) \cap (\vec{X_c}=\vec{x_c})$, where $\vec{X_b}$ gives the set of intermediate nodes of the causal chain of the counterfactual action $b$, and $\vec{X_c}$ is generated using the Algorithm \ref{alg:contrastive}. Values $\vec{r}$, $\vec{x_c}$ are contrasted using the actual instantiation $M_{\vec{\mathcal{V}}\leftarrow \vec{S}}$ and counterfactual instantiation $M_{\vec{Z}\leftarrow \vec{S_Z}}$ from Definition \ref{def:actual_instantiation}.
\hfill\BlackBox 
\end{definition}

\tm{As with earlier definitions, a bit of explanation would make this part clearer}
\pmm{added below}

Instead of just having the intermediate nodes of the causal chain of the \emph{actual} action (as in Definition \ref{def_why_dt}), we now get the set of intermediate nodes for for the \emph{counterfactual} action from its causal chain. Then the intermediate nodes of the counterfactual chain is compared with the set of nodes we get from the Algorithm \ref{alg:contrastive}, to get the common set of nodes, of which the variable values will finally be contrasted. 

As before, we explain Definition \ref{def_whynot_dt} using the adversarial StarCraft II task. Consider the question `Why not action $A_b$', when the actual action is $A_s$, for which the explanation is generated as follows. We first get the decision nodes $A_n$ and $B$ having $<=5$ and $>2$ as the decision boundaries respectively. Then each decision boundary value starting with the node closest to the leaf node, is moved by a small $\Delta$ amount $0.01$ and applied as the new feature value in the current state of the agent ($B$ feature value will change to $1.99$). We use this new state to predict the counterfactual action as $A_b$ from the decision tree, and to get the counterfactual decision nodes (which remains the same). Next, we get the intersection of nodes in the causal chain of the counterfactual action $A_b$ ($B \rightarrow A_n \rightarrow [D_u, D_b]$) with  
$\vec{X_c}$, which gives $B$ as $\vec{X_{con}}$ with the actual value $3$ and counterfactual value $1.99$. Finally, these values are contrasted and appended with the reward nodes of the causal chain of $A_b$ to generate the explanation. A graphical interpretation of this explanation is shown in  Figure \ref{fig:contrast_example}.

\tm{This above example is a nice example, but perhaps needs a bit of graphical representation, or at least, a cleaner representation; e.g. show the nodes and variables for 'why', show them for 'why not', show show to integrate the action influence graph, and then show the final explanation}

\pmm{showed in a graphical way below}

\begin{figure*}[!t]
        \centering
        \resizebox{0.8\textwidth}{!}{%



        }
    \caption{Visual example of the explanation for \emph{Why not $A_b$}. Actual action and the actual causal chain is shown in blue, counterfactual chain and nodes are shown in red and the contrast node is shown in green.}
    \label{fig:contrast_example}
    \end{figure*}

\subsection{Learning Opportunity Chains}

\tm{Briefly motivate this (again) by outlining the weakness of just explaining the policy alone: it doesn't consider that actions are chosen to enable other actions}

\pmm{added some sentences below.}

Explaining the behaviour of the agent using only the policy (or a surrogate policy) alone, even if the explanation is causal, has shortcomings as this does not consider that some actions might be chosen because they enable other actions. In this section we discuss how information on enabling actions can be used to form a more complete explanation.

In the context of reinforcement learning, we define a `distal action' as the action that depends the most on the execution of the \emph{current} action of the agent. The agent might not be able to execute the distal action unless some other action was executed first (i.e. \emph{some} actions `enable' the execution of other actions). For example, in the StarCraft II domain, the action `train marines' cannot be executed until `build barracks' action is executed. While it is possible to extract distal actions from environment dynamics and pre-conditions in a model-based system, for model-free RL agents, this remains a challenge. However, for the purpose of explanation, it is possible to provide an approximation and predict the distal action.  

\tm{I think we need a problem formulation here: what are the inputs and outputs of the prediction? During training, what is the action that is used as the ground truth; just the next action? etc. Define the learning problem a bit more formally. As it stands, I would struggle to re-produce this}
\pmm{I have explained this informally in below but added a formal defintion as well to be clear and the ground truth}

We use a many-to-one recurrent neural network (RNN)~\citep{schuster1997bidirectional} as our prediction model $\mathbb{\widehat{L}}$ to approximate the distal action given a sequence of previous states and actions of the agent. We implement $\mathbb{\widehat{L}}$ with a fully connected hidden layer of 10 units, and a batch size of 100. For training data, we use the batch replay dataset $D_t$ discussed in Section 4.2. We define a sequence as a state-action trace that ends in an action that is one of the last actions in a causal chain (e.g. in Figure 2, the last action of all causal chains in the `attack' action). The output of the model $\mathbb{\widehat{L}}$ will be the distal action and its expected cumulative reward. Formally, this action prediction model can be written as: $\hat{y}_{t_{N+1}} = f(x_{t_1}, x_{t_2}, ... , x_{t_N};t_1, t_2, ..., t_N )$, where $x_{t_N}$ is the state-action pair (including state features) of the last action of a causal chain, $\hat{y}_{t_{N+1}}$ gives the distal action and the reward. Here, we use the immediate next action that lies in a particular causal chain as the ground truth. Note that even though we used an RNN to implement the prediction model, it is entirely possible to use other models to approximate the distal action. With the distal action prediction model $\mathbb{\widehat{L}}$ in hand, we now define \emph{minimally complete distal} explanations for `why' and `why not' questions that incorporate causal nature to the explanations.     

\begin{definition}
\label{def_distal}
Given a \emph{minimally complete contrastive} explanation, current action $a$ and a prediction model $\mathbb{\widehat{L}}$, a \emph{minimally complete distal} explanation is a tuple $\left ( \vec{R}=\vec{r}, \vec{X_{con}}=\vec{x_{con}}, a_{d}\right )$, in which $\vec{R}$ and $\vec{X_{con}}$ do not change from Definition \ref{def_whynot_dt}; and $a_{d}$ gives the distal action predicted through $\mathbb{\widehat{L}}$ such that $a_{d} \in A \cap A_c$, where $A$ is the action set of the agent and $A_c$ gives the action set of the causal chain of current action $a$. 
\hfill\BlackBox
\end{definition}

\tm{In the paragraph below (and in fact where we use the variable names from the graph), I suggest expanding the action/variable names -- it is easy to forget what $A_b$ means}
\pmm{done}
Informally, this simply prepends the predicted distal action to a minimally complete contrastive explanation generated through Definition \ref{def_whynot_dt} if the distal action exists in the causal chain of the current action. Consider the example `Why not action \emph{build\_barracks} ($A_b$) , when the actual action is \emph{train\_marine} ($A_m$). This would yield the counterfactual decision node $A_n$ (ally unit number) with the actual value $10$ and the counterfactual value $5$. When the predicted distal action is \emph{attack} ($A_a$), we can generate the below explanation text using a simple natural language template. The causal explanation is generated with Definition \ref{ation_influence_minally_complete} while the distal explanation is generated through Definition \ref{def_distal}.

\liz{I confess to this example confusing me by the terminology - here $A_a$ seems like a `proximal' (or even `immediate') rather than a `distal' action - using my understanding of the terminology from \cite{hilton2005course}}
\pmm{Its true that for Hilton 2005, the distal cause would far out cause in the causal chain because in that case the explanation is about past events only. In our case explanation is about the behaviour of the agent in the future as it involves rewards so the distal cause/action becomes an action far out in the future but still influenced by present actions/events. i added a small description of this for the discussion in the first study}

\begin{center}
\noindent
\begin{tabular}{@{}l@{~~}p{0.6\textwidth}@{}}

\emph{Causal Explanation: } & Because it is more desirable to do the action train marine ($A_m$) to have more ally units ($A_n$) as the goal is to have more Destroyed Units ($D_u$) and Destroyed buildings ($D_b$).\\

\emph{Distal Explanation: } & Because ally unit number ($A_n$) is less than the optimal number 18, it is more desirable do the action train marine ($A_m$) to \emph{enable the action} attack ($A_a$) as the goal is to have more Destroyed Units ($D_u$) and Destroyed buildings ($D_b$).
\end{tabular}
\end{center}

\tm{In the above example, where did the `more desirable to do the action training marine' go? Why isn't this included in the distal explanation? Similarly, why does the causal explanation have the `optimal number 18'? This comes from the action influence graph right?}
\pmm{more desirable is a filler text, which is also present in the distal explanation in second sentence. value is coming from the decision boundary}

Note that the Definition \ref{def_distal} can also be used in conjunction with the Definition \ref{def_why_dt} to generate distal explanations for `why' questions.

\subsection{Computational Evaluation}

\begin{table}
\centering
\small
  \begin{tabular}{p{3.2cm}@{ }r r r @{ }r c @{ }c}
    \toprule
    \multirow{2}{*}{Env - RL} &
      \multicolumn{1}{c}{} &
      \multicolumn{3}{c}{SE - Accuracy (\%)} &
      \multicolumn{2}{c}{DP - Accuracy (\%)} \\
      \cmidrule(lr){3-5}\cmidrule(l){6-7}
      & {Size} & {LR} & {DT} & {MLP} & {$DP$} & {$DP_n$} \\
      \midrule
    Cartpole-PG  & 4/2 &     83.8 & 81.6 & 86.0      & 96.83 & 97.10  \\
    MountainCar-DQN & 3/3 &    69.7 & 57.8 & 69.6     & 88.66 & 86.75  \\
    Taxi-SARSA &  4/6 &       68.2 & 74.2 & 67.9       & 82.44 &  86.19 \\
    LunarLander-DDQN  & 8/4     & 68.4 & 63.7 & 72.1      & 72.82 &  72.91  \\
    BipedalWalker-PPO  & 14/4     & 56.9 & 56.4 & 56.7   & 67.99 &  69.28\\
    StarCraft-A3C  & 9/4       & 94.7 & 91.8 & 91.4       & 97.36 & 86.04 \\

    \bottomrule
  \end{tabular}
 
  \caption{Distal explanation model evaluation in 6 benchmark reinforcement learning domains that use different RL algorithms, measuring mean task prediction accuracy in 100 episodes after training. SE-structural equations (trained with LR-linear regression, DT-decision trees, MLP-multi layer perceptrons), $DP$-decision policy tree and $DP_n$-unconstrained decision policy tree.}
  \label{tablecompute}
\end{table}

 \tm{The bold numbers in Table~\ref{tablecompute} could be a bit misleading. Bold usually means ``best'', but the $DP_n$ model is better in four out of six cases}
 \pmm{removed}
 
\tm{We use task prediction as a measure of faithfulness, but I wonder if we shouldn't also measure 'effectiveness'. That is, use the predicted action as the next action to see how `good' our predictions other in the problem. For example, the surrogate model may be 80\% faithful, but if we use it to `play' the benchmarks 100 episodes each, what is the average reward over those 100 episodes. If it is similar to the underlying black-box policy, you could argue to \emph{replace} the black-box policy all together and just use the surrogate policy, then the surrogate policy is 100\% faithful to itself}
 
We use five OpenAI benchmarks~\citep{openai} and the adversarial StarCraft II scenario (discussed in Section 5.1) to evaluate the task prediction~\citep{hoffman2018metrics} accuracy of our distal explanation model and compare against action influence models \citep{madumal2019explainable} as a baseline. Task prediction can be used to predict what the agent will do in the next instance, and measures how faithful the surrogate policy is against the  underlying policy.

We choose the benchmarks to have a mix of complexity levels and causal graph sizes (given by the number of actions and state variables). We train the RL agents using different types of model-free RL algorithms (see Table \ref{tablecompute}), using a high performance computer cluster node with 2 Nvidia V100 GPUs, 56GB of memory and 20 core CPU with 2.2GHz speed. All agents were trained until the reward threshold (to consider as `solved') of the environment specification is reached.

We evaluate two versions of the distal explanation model, where one is a based on a depth limited decision tree with the number of actions ($DP$ in table \ref{tablecompute}), other trained until all leaves are pure nodes ($DP_n$). Results summarised in Table \ref{tablecompute} show our model outperforms task prediction of action influence models (with their structural equations trained by either linear regression (LR), decision trees (DT) or multi layer perceptrons (MLP)) in every benchmark, some by a substantial margin.  

The benefit gained through unconstrained decision trees ($DP_n$) does not translate well into an increase in task prediction accuracy. We conclude that for the purpose of using distal models for explanation, a depth limited tree ($DP$) provide an adequate level of accuracy. Moreover, as a depth limited tree is likely to be more interpretable to a human, it is more suited for \emph{explainability} and \emph{explanation}.

\section{Evaluation: Human Study}

We consider human subject experiments to be an integral part of XAI model evaluation and as such conduct a human study with \textbf{90} participants. We consider two hypotheses for our empirical evaluation; 1) Distal explanation models leads to a improved \emph{understanding} of the agent; and 2) Distal explanation models provide \emph{subjectively} `better' explanations. Our experiment involves RL agents that complete objectives in three distinct scenarios, which are based on the StarCraft II~\citep{vinyals2017StarCraft} learning environment. We first discuss these scenarios below. 

\subsection{Scenarios}
\label{section5-1}
\liz{consider presenting the scenarios early in the paper to provide motivation; can then use as `running' example}
\pmm{not sure if this should be added to the introduction or somewhere else as it is a bit wordy}

\begin{figure}[!t]
\includegraphics[width=\columnwidth]{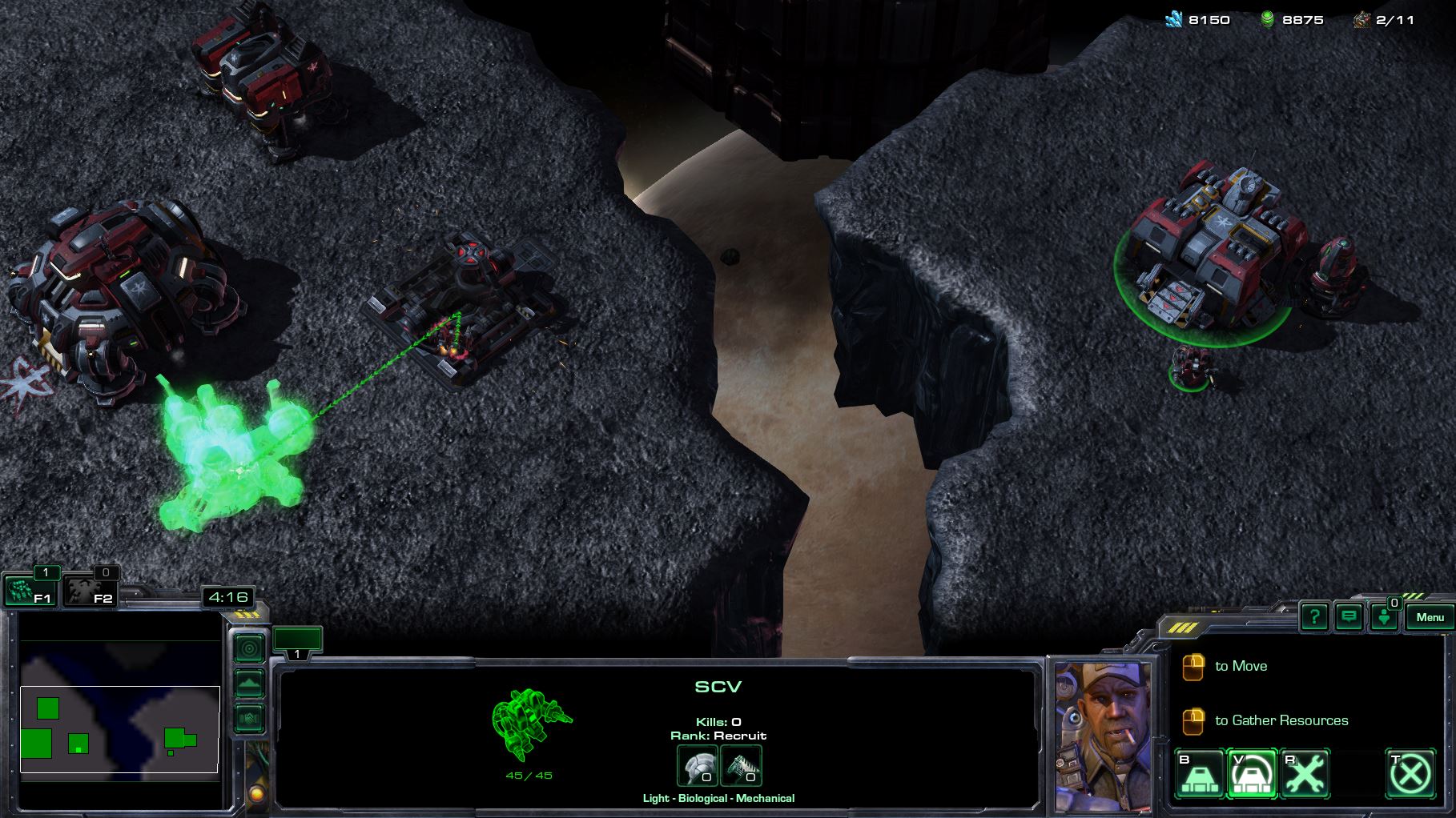}
\caption{StarCraft II Collaborative task scenario: The agent is controlling the leftmost section and the participant controls the right section (divided by the fissure)} \label{fig_collab}
\end{figure}

\tm{As noted earlier, I would suggest making is clear that the other scenarios are part of the general starcraft game -- we just use the simulation frameowkr to generate additional scenarios}
\pmm{done}

In addition to the default scenario of the StarCraft II, we developed two additional scenarios as custom maps using the StarCraft II platform as a framework, that are better suited for explainability. Custom maps were made to add a more strategic nature to scenarios and in some cases to elicit cooperation from the interacting human. Note that these scenarios are completely different from the StarCraft II \emph{game}. We only use StarCraft II assets as a simulation framework, similar to how e.g.\ a grid-world framework can be used to make many different scenarios. We also release these maps with state and action specifications as test-beds for explainability research.

\liz{justify `better suited'}
\pmm{added s small justification}

\paragraph{Adversarial} In this scenario, the agent's objective is to build its base by gathering resources and destroy the enemy's base. The agent can build offensive units (marines) to attack the enemy's base and to defend its own base. This is the default objective in a normal StarCraft II game, but here we only use 4 actions for the purpose of the experiment rewards are given for the number of enemies and buildings destroyed (shown in Figure \ref{fig:causal_nodes} b) as an action influence graph). During the experiment, the trained RL agent will provide explanations to the participant and the strength of the explanations are evaluated through task prediction.

\paragraph{Rescue} This scenario is a custom map, where the agent's objective is to find a missing unit and bring it back to the base using an aerial vehicle. The agent also has to avoid or destroy enemy units during the rescue and aid the aerial vehicle using an armed unit. The agent has access to 5 actions, the reward is given for the number of missing units saved. The evaluation is done through task prediction as before.

\paragraph{Collaborative Task} The collaborative task is fundamentally different from the previous scenarios, in that the participant has to help the agent to complete the objective. We made this task as a custom map (depicted in Figure \ref{fig_collab}) where the map is partitioned as the agent and human `area'. The agent can perform 5 actions in this task, while the human can choose 4 actions to execute. The objective of the task is to build a series of structures that finally leads to the creation of an `elite' unit, which the human has to transport to a base. The success of the task depends on the participant choosing to execute the action that best support the agent.

\subsection{Experiment Design and Methodology}

\tm{Isn't the experiment designed mixed? We use within subject across explanation types, and between subject to compare across scenarios?}
\pmm{changed}

To investigate the two main hypotheses, we use a mixed design~\citep{keren2014between} (within subject and between subject) for our experiment. Every participant will be evaluated on the 3 independent variables which are 1) `no explanations', where only a visual description of the agent behaviour is provided; 2) causal explanations generated with action influence models~\citep{madumal2019explainable} and 3) our distal explanation model. At a glance, the experiment has 3 phases where participants receive explanations from RL agents, subjectively evaluate the explanation and are then evaluated through task prediction~\citep{hoffman2018metrics} to gauge their understanding of the agent.

Task prediction is an effective measure that can peek into the mental model of an explainee to evaluate how successful the given explanation was in transferring the knowledge from the explainer~\citep{hoffman2018metrics,miller2018explanation}. In task prediction, the participant is asked the question `What will the agent do next?'. We use task prediction to evaluate the hypothesis 1) for the Adversarial and Rescue scenarios, and invert the question as to ask `What would you do next?' in the Collaborative task. We investigate hypothesis 1) by employing the 5-point Likert \emph{explanation satisfaction scale} of ~\citeauthor{hoffman2018metrics}~(\citeyear[p.39]{hoffman2018metrics}). Explanation satisfaction is evaluated after each explanation and also at the end of the experiment which compares explanations of causal and distal models.

\noindent\textbf{Experiment Design:} We use \emph{Amazon Mechanical Turk} (Mturk)---a crowd sourcing platform well known for obtaining human-subject data ~\citep{buhrmester2011amazon}---to conduct the experiments. A web-based interactive interface is used as the medium of interaction.

\begin{figure}[!t]
\includegraphics[width=\columnwidth]{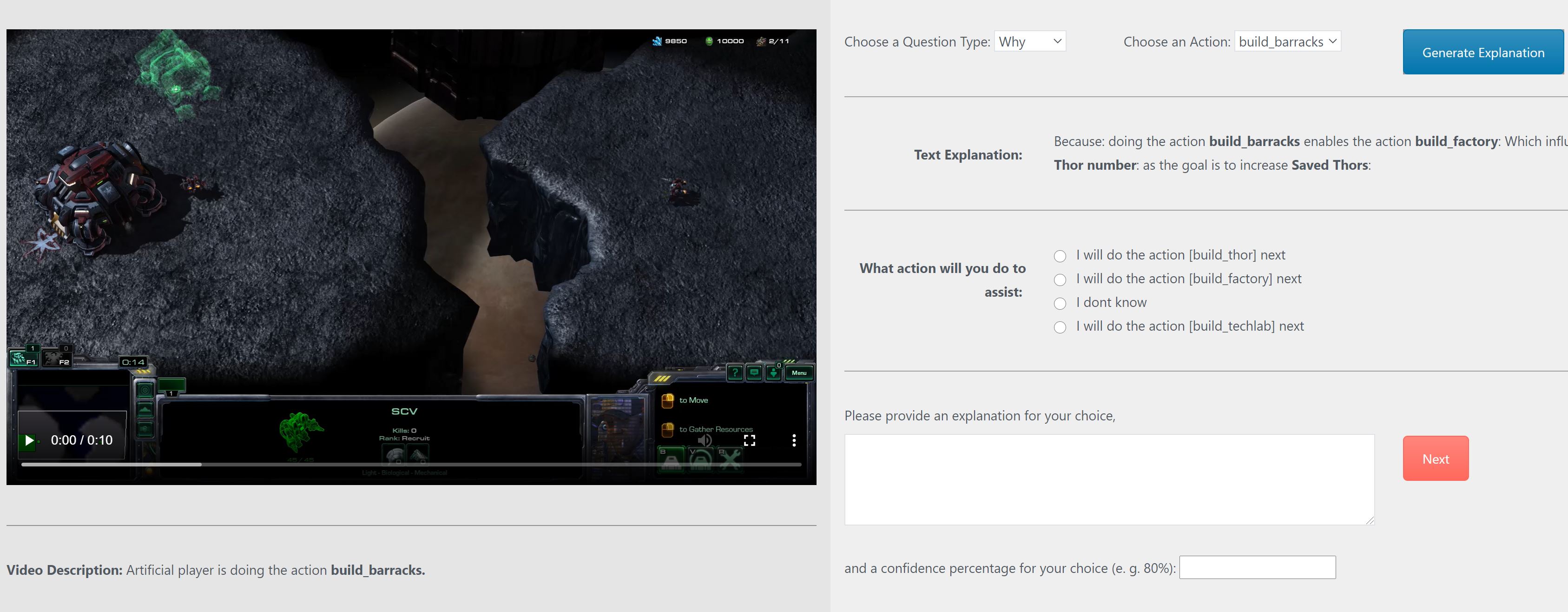}
\caption{The web-based interface of the experiment showing the Collabrative Task.} \label{fig_interface}
\end{figure}

We first display the ethics approval obtained through a university, and after the participants' consent gather demographic information. We then show video clips of the agents solving the 3 StarCraft II scenarios that capture the behaviour of the agents. Each scenario has 4 distinct behaviours of the respective agent (around 10 seconds per clip).  Every participant sees all three scenarios, and all three explanation types, but between participants, the combination of scenario and explanation type are mixed. For example, a participant may experience: Adversarial with no explanations, Rescue with casual explanations and Collaborative with distal explanations. The order of these is randomised to control for ordering effects.

The first stage of the experiment involves training the participants to identify agents' actions using video clips of the agents performing those actions before the start of each scenario. In the Collaborative scenario, participants are trained to identify the actions they can use instead. After validating that participants can distinguish different actions through a question, the scenario will be presented.

The second stage lets the participants ask explanatory questions (in the form of why/why not \emph{action}), after watching the agent's behaviour through the video clip. Participants can ask any number of questions and we did not control for a minimum number of questions; however,  we incentivised participants to ask questions because they knew they would receive bonus payments for getting predictions correct later in the experimence. After each explanation video, participants are presented with the explanation satisfaction survey. For each explanation/scenario pair, each participant engages in 4 tasks.

\tm{Below: ``This process is also repeated 4 times.'' Not clear whether this is 5 in total? Once then 4 repeats. Or whether just 4. I think it is the latter (if I recall correctly), so re-word to something like: ``Each participants makes predictions for 4 tasks''}
\pmm{changed}

The third stage involves evaluating the participants' `understanding' of the agent through task prediction. Participants are presented with 4 new videos with different situations, and are asked what action the agent will do next, and can select one of the 4 options (which are 3 actions of the agent plus the option of `I don't know'). Each participants makes predictions for 4 tasks. After this stage participant will move to the next scenario with a different explanation model and repeat from Stage 1 to 3. This is done until all the scenarios are encountered by the participant.

In the final stage, the participant is presented with 3 additional explanation videos (of the scenario they did for the no explanation condition), and is presented with causal explanations from action influence models~\citep{madumal2019explainable} and our distal explanation model \emph{side by side}. We use ~\citeauthor{hoffman2018metrics}~\cite[p.39]{hoffman2018metrics}'s explanation satisfaction scale but this time as a movable slider that subjectively compares the two explanation.

\noindent\textbf{Experimental Conditions: }
We ran the experiment with the above mentioned 3 independent variables (the explanation models), which resulted in 3 combinations of explanation model and scenarios, with participants seeing all 3 scenarios and all 3 explanation types. Each combination had 30 participants for a total of 90 participants in the experiment. Each participant is scored on the total number of correct task predictions out of 12 (4 each for each model-scenario combination).

Each experiment ran approximately 50 minutes, and we compensated each participant with 8.5USD (a bonus compensation of 0.5USD was also given to participants for each point above 10). Participants were aged between 23 to 60 ($\mu=38.1$), and of the 90 participants, 51 were male while 38 were female and 1 who did not provide an answer. Participants reported an average self-rated gaming experience and StarCraft II experience of 2.47 and 1.47 out of 5 (5-point Likert) respectively.

\tm{A few bits of information missing below: ``If the participant failed this, the experimence did not proceed" -- I think they were still paid some amount? Just state this and state how much. Also state the threshold use. And also not how many participants were excluded}
\pmm{added some text}

To ensure the quality of data from participants, we  recruited only `master class' workers with $95\%$ or more approval rate. We controlled for language by only recruiting workers from the United States. We excluded the noisy data of users in 3 ways. First, we tested participants to ensure they had learnt about the scenario by asking them to identify actions shown in several videos. If the participant failed this, the experiment did not proceed (participants were paid a \$2USD base amount). Second, we tracked how much time each participant spent viewing explanations and answering tasks. If this was regularly below a threshold of a few seconds, we omitted that participant from our results. Third, participants were required to explain their task predictions. If this text was gibberish or a 1-2 word response, we omitted that participant from the results. We filtered out 16 participants according to the above constraints to yield the final participant number of 90.

\subsection{Results}

\begin{figure}[!ht]
\centering
\includegraphics[width=0.7\columnwidth]{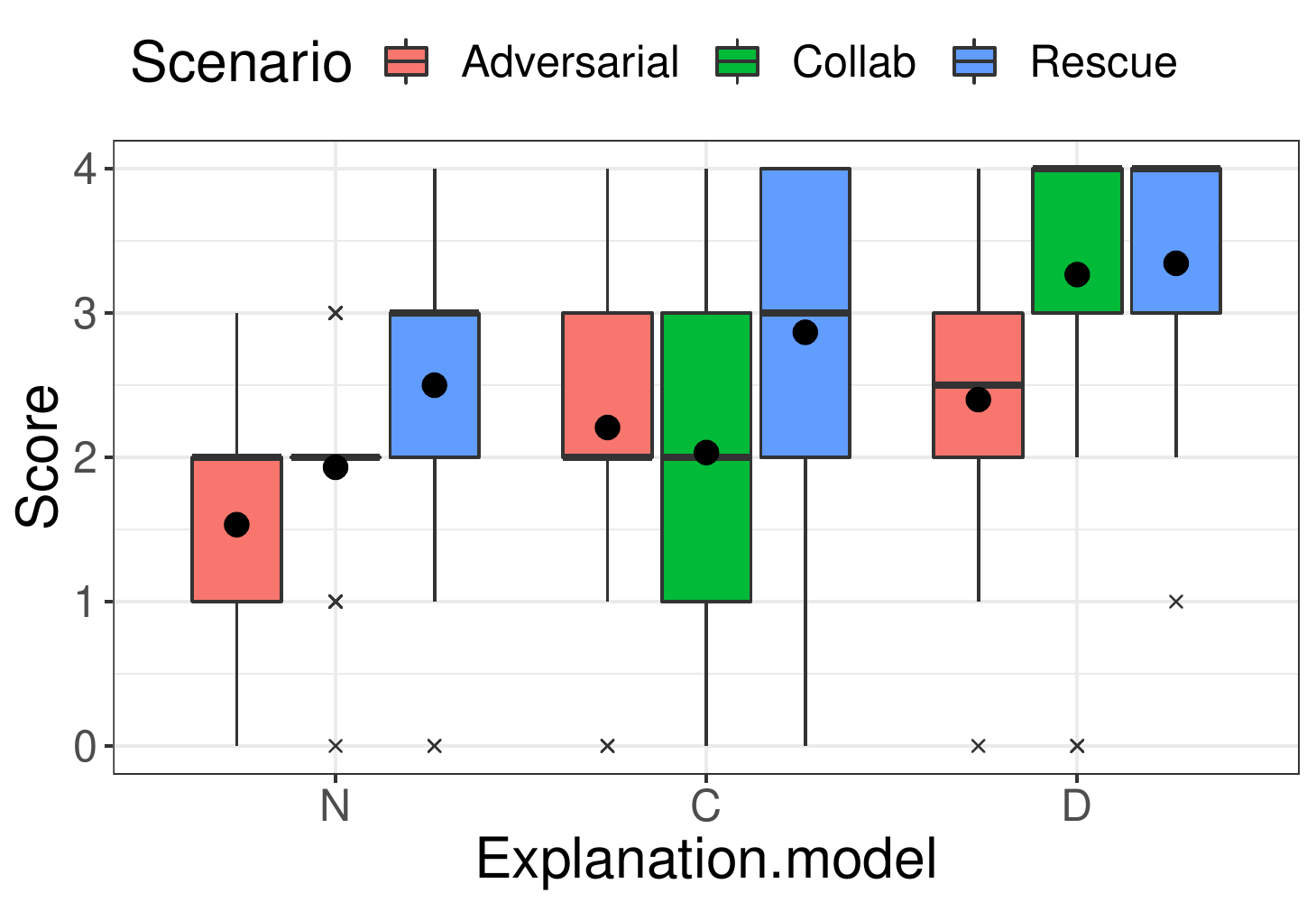}
\caption{Box plot of task prediction scores of the explanation models across the StarCraft II scenarios (means are represented by bold dots)} \label{fig-scores}
\end{figure}

\begin{table}
\centering
\caption{Pairwise differences with a z-test for proportions for each model-pair and pairwise t-tests in the three StarCraft II scenarios in task prediction scores, considering the correct response.}
\label{pairwise-taskprediction}

\begin{tabular}{p{3cm}@{ }l r r r r r}

\toprule

\multirow{2}{*}{Model ($m_1$-$m_2$)} &  
      \multicolumn{1}{c}{} &
      \multicolumn{3}{c}{Z-test} &
      \multicolumn{2}{c}{T-test} \\
      \cmidrule(lr){3-5}\cmidrule(l){6-7}
      & {Scenario} & {$X^2$} & {p-value} & {Prop ($m_1$|$m2$)} & {p-value} & {t-stat} \\
      \midrule


\multirow{3}{*}{C - N}
    & Adversarial & 5.437  & 0.019 & 0.53 | 0.38 & 0.056 & -1.988\\
   & Rescue  & 2.283 &  0.130 & 0.71 | 0.62 & 0.169 & -1.408\\
   & Collaborative  & 0.416 & 0.518 & 0.50 | 0.46 & 0.537 & -0.623\\
   \hline
  
\multirow{3}{*}{D - N}
    & Adversarial &  11.269  &  \textbf{<0.001} & \textbf{0.60 | 0.38} & \textbf{<0.001} & \textbf{-3.791}\\
   &  Rescue  & 9.931 &   \textbf{0.001} & \textbf{0.80 | 0.62} & \textbf{0.010} & \textbf{-2.750}\\
   & Collaborative  & 31.966 &  \textbf{<0.001} & \textbf{0.81 | 0.46} & \textbf{<0.001} & \textbf{-4.761}\\
\hline
\multirow{3}{*}{D - C} 
    & Adversarial & 1.085  &  0.297 & 0.60 | 0.50 & 0.325 & -1.000\\
   & Rescue  &  2.784 &   0.095 & 0.80 | 0.71 & 0.221 & -1.249\\
   & Collaborative  & 25.511 & \textbf{<0.001} & \textbf{0.81 | 0.50} & \textbf{<0.001} & \textbf{-4.367}\\
\bottomrule

\end{tabular}

\end{table}

We first discuss the results on hypothesis 1), where we investigate whether distal explanation models lead to a better understanding of the agent. We present the null hypotheses as $H_0: P_N = P_C = P_D$ and the alternate hypothesis as $H_1 : P_D >  P_N$ and $H_2 : P_D >  P_C$, in which N, C, D corresponds to `no explanation', causal and distal explanation models. Here, $P$ denotes the proportions of the observed values of correct answers in task predictions by the participants.

We perform Pearson's Chi-squared test for the three StarCraft II scenarios and obtain the following values: Adversarial (p-value = 0.011, $X^2 = 13.00$), Rescue (p-value = 0.034, $X^2 = 10.40$) and Collaborative (p-value = <0.001, $X^2 = 35.47$). As the Chi-squared test was significant at the 0.05 level across the three scenarios, we investigate the pairwise differences between models using a z-test. We summarise the results in Table \ref{pairwise-taskprediction}. From Table \ref{pairwise-taskprediction}, considering the proportions ($P$) between model pairs, we can see that apart from Adversarial and Rescue scenarios for the D - C model pair, distal explanation models have statistically significant results at the 0.05 level between other combinations. Thus we accept $H_1$ for every StarCraft II scenario and accept $H_2$ only for the Collaborative scenario. We further test the validity of our results by employing a pairwise t-test which produce similar conclusions (results shown in Table \ref{tablecompute}. We illustrate these results as a box-plot in Figure \ref{fig-scores}. Clearly, the Collaborative scenario poses a much higher challenge to the participants, and results indicate that distal explanations perform better than other models in this task.

\begin{table}[!t]
\caption{Pairwise differences with a z-test for \emph{explanation quality} metrics in models Distal (D) vs Causal (C), data where participants rated `5'.}
\label{pairwise-explanation-quality}
\centering
\begin{tabular}{llrrr}

\toprule

Metric  & Scenario & $X^2$ & p-value & Proportions (D|C) \\\midrule

\multirow{3}{*}{Complete}
    & Adversarial  & 3.267 & 0.070& 0.56 | 0.45\\
   & Rescue  & 0.074 &  0.785 & 0.33 | 0.35\\
   & Collaborative & 11.428 & \textbf{<0.001}& \textbf{0.40 | 0.20}\\
   \hline
  
\multirow{3}{*}{Sufficient}
    & Adversarial  &  6.020 & \textbf{0.014}& \textbf{0.56 | 0.40}\\
   & Rescue  & 0.018 &  0.892 & 0.35 | 0.34\\
   & Collaborative & 15.55 & \textbf{<0.001}& \textbf{0.41 | 0.18}\\
\hline

\multirow{3}{*}{Satisfying}
    & Adversarial  & 1.085 & 0.297& 0.46 | 0.40\\
   & Rescue  & 1.528 &  0.216 & 0.29 | 0.36\\
   & Collaborative & 9.981 & \textbf{0.001}& \textbf{0.42 | 0.23}\\
\hline

\multirow{3}{*}{Understanding}
    & Adversarial  & 1.377 & 0.240& 0.46 | 0.39\\
   & Rescue  & 0.071 &  0.788 & 0.35 | 0.37\\
   & Collaborative & 15.31 & \textbf{<0.001}& \textbf{0.42 | 0.19}\\
\bottomrule

\end{tabular}

\end{table}

\begin{figure}[!ht]
\centering
\includegraphics[width=\columnwidth]{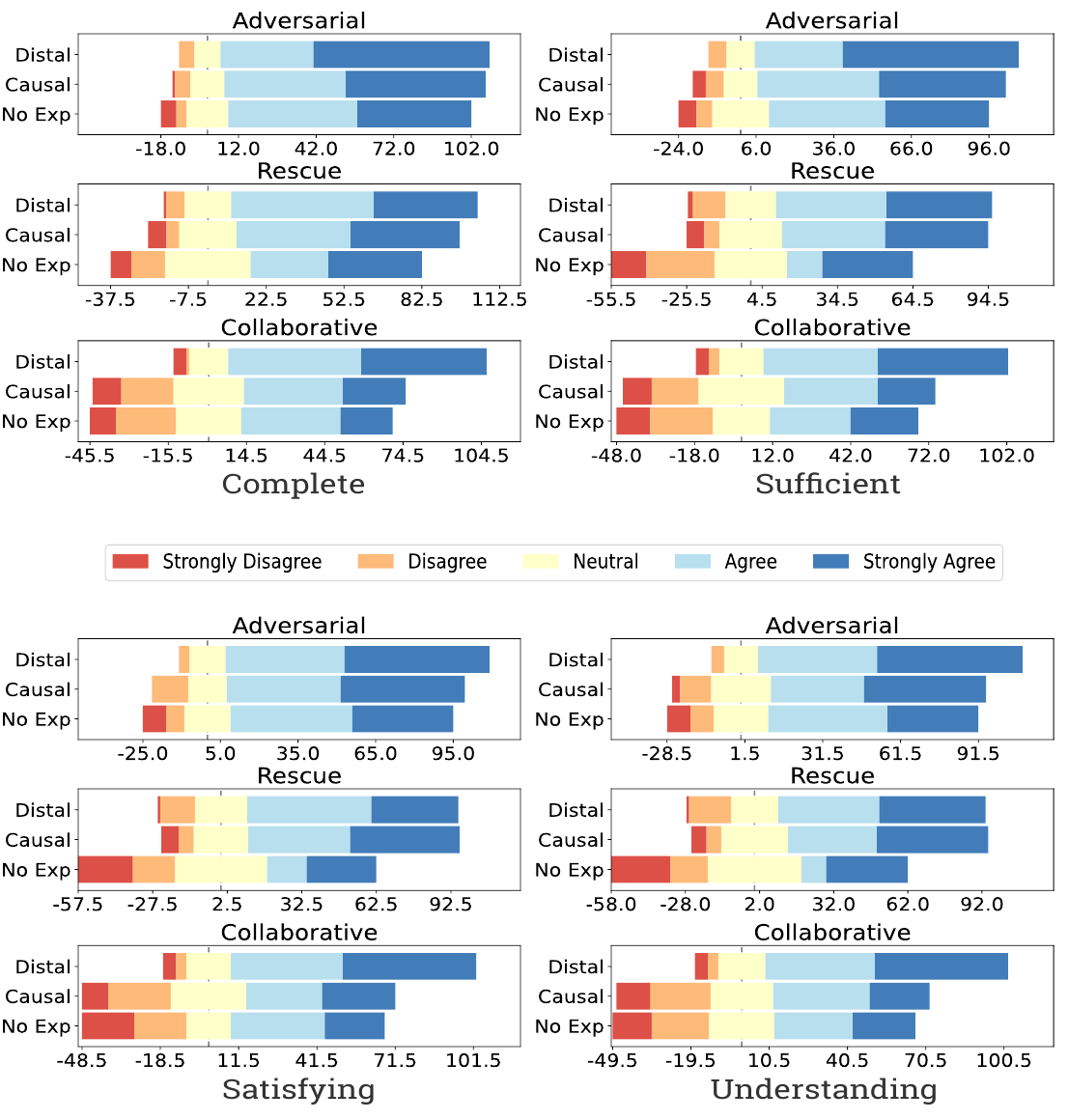}
\caption{Likert scale counts of explanation quality metrics and how they vary across explanation models and scenarios. X-axis represent the total counts each Likert category received, adjusted to represent 0 as the midpoint.} \label{fig_quality}
\end{figure}

\tm{The results use $H_2$ for statistical testing and H2 for the experimental hypothesis, which is a bit confusing. Perhaps don't use H1 and H2 for the experimental hypothesis; just use 1 and 2?}
\pmm{changed}

\textbf{Explanation Quality: } The second main hypothesis 2), evaluate whether distal explanations can provide \emph{subjectively} better explanations. The corresponding null hypothesis is $H_0: P_N = P_C = P_D$ and the alternative hypothesis is $H_1 : P_D >  P_C$. $P$ in this case $P$ becomes the proportion of the observed values of the Likert scale data (using the survey of ~\citeauthor{hoffman2018metrics}~(\citeyear[p.39]{hoffman2018metrics})), where participants have rated as `5'. We consider four explanation quality metrics; `Complete', `Sufficient', `Satisfying' and `Understanding'. As before, we employ Pearson's Chi-squared test to see the significance of the above 4 metrics in the 3 StarCraft II scenarios, and obtain p-values < 0.01 for every condition. As there are significant differences between explanation models on explanation quality, we reject $H_0$ and conduct a pairwise z-test. We summarise the results in Table \ref{pairwise-explanation-quality}. Figure \ref{fig_quality} captures the Likert scale data distribution across models and scenarios. Though it is visually evident that distal explanation quality across the compared metrics has a positive trend, from the three scenarios, only the Collaborative task yields significant results for every explanation quality metric (see Table \ref{pairwise-explanation-quality}). Thus we accept $H_1$ only for the Collaborative scenario.

\tm{What is the x-axis measure for the likert scales? 0 is neutral, 100 is everyone rating 5, and -100 is everyone rating 1? Needs some explanation}
\pmm{added a description to the figure title}

\paragraph{Discussion: } The results we obtained for \emph{explanation quality} mirror the results in task prediction. Intuitively this makes sense as participants are more inclined to rate an explanation `good' if they feel they have a better `understanding' of the agent. Further investigations are needed to explore why distal explanations perform substantially better in human-agent collaborative tasks. 

To investigate whether the knowledge of the StarCraft II game had any impact on task prediction scores, we perform a Pearson's correlation test between task prediction and StarCraft II experience (self-report in a 5-point Likert scale). The obtained values (t = 1.515, p-value = 0.133) indicate that there is no statistically significant correlation between scores and StarCraft II experience. Although our experiment was based on the StarCraft II environment, we used custom maps and scenarios that are different from the game. Thus the results of the correlation test is plausible.   

One weakness of our model is the need for a causal graph that is faithful to the problem, in order to learn the \emph{opportunity chains}. For the purpose of this paper, we hand-crafted the causal graphs for StarCraft II scenarios and the 5 RL benchmarks. While our hand-crafted models can be verified easily with data, we acknowledge that it may become infeasible in larger domains. We view generating a causal graph a distinct problem than generating explanations \emph{using} a causal graph. As such we propose this as our immediate future work.  
 
 \tm{Here we need a section on the limitations of the experiment design: external and internal validity in particular}
 \pmm{added some explanation}
 
 \paragraph{Limitations of the experiment design: } Although we used scenarios that have different objectives than the standard Starcraft II game, familiarity with the game's concepts may have had some impact on the scores even if it is not significant. Participants also may have had commonsense knowledge about such scenarios (in particular rescue and adversarial) that can affect their judgments. Our results should be generalisable across similar scenarios in different \emph{domains}, though further experiments are needed to evaluate the generalisability across different scenarios (e.g. path planning, manufacturing).

\section{Conclusion}

We introduce a \emph{distal explanation} model for model-free reinforcement learning agents that can generate explanations for `why' and `why not' questions. These models learn \emph{opportunity chains} (in the form of $A$ enables $B$ and $B$ causes $C$), and approximate a future action that \emph{enables} due to the current action of the agent. Our motivation comes from insights gained through a human-agent experiment, in which we analysed \textbf{240} human explanations. Participants in this study frequently referred to future action that depend on the current action of the agent, which conform to the definition of opportunity chains. To learn opportunity chains at the training phase in reinforcement learning we make use of \emph{action influence models} to extract causal chains and represent the approximated policy of the agent in a decision tree policy. In contrast to action influence models that use structural causal equations to generate \emph{contrastive} explanations, we use the decision policy in conjunction with causal chains to improve the accuracy of \emph{task prediction}. We evaluate our approach in 6 RL benchmarks on task prediction. We then undertake a human study with \textbf{90} participants to investigate how the distal explanation model perform in task prediction and \emph{explanation quality} metrics in three custom scenarios built using the StarCraft II platform.

While results indicate a significantly better performance of distal explanations compared with two other explanation models in collaborative situations, further research is needed to understand the impact this technique may have on other types of scenarios. One weakness of our model is the need of knowing the causal structure of the domain beforehand. Though this can be mitigated by using existing causal discovery methods, the reinforcement learning setting provides a unique opportunity to learn causal graphs that are better suited for explanation through influencing the exploration of the agent. Thus our immediate future work involves generating causal models at the training time of the agent through causal discovery.    

\tm{Suggest that you go through the references carefully to look for typos etc; e.g. in Brockman ``Openai'' should be ``OpenAI'', and is one Elizalbe reference, ``mdp'' should be ``MDP''. Probably there are others}

\pmm{fixed}








\vskip 0.2in
\bibliography{sample}

\begin{thebibliography}{47}
\providecommand{\natexlab}[1]{#1}
\providecommand{\url}[1]{\texttt{#1}}
\expandafter\ifx\csname urlstyle\endcsname\relax
  \providecommand{\doi}[1]{doi: #1}\else
  \providecommand{\doi}{doi: \begingroup \urlstyle{rm}\Url}\fi

\bibitem[Abdul et~al.(2018)Abdul, Vermeulen, Wang, Lim, and
  Kankanhalli]{abdul2018trends}
Ashraf Abdul, Jo~Vermeulen, Danding Wang, Brian~Y Lim, and Mohan Kankanhalli.
\newblock Trends and trajectories for explainable, accountable and intelligible
  systems: An {HCI} research agenda.
\newblock In \emph{Proceedings of the 2018 CHI conference on human factors in
  computing systems}, page 582. ACM, 2018.

\bibitem[Amir and Amir(2018)]{amir2018highlights}
Dan Amir and Ofra Amir.
\newblock Highlights: Summarizing agent behavior to people.
\newblock In \emph{Proc. of the 17th International conference on Autonomous
  Agents and Multi-Agent Systems (AAMAS)}, 2018.

\bibitem[B{\"o}hm and Pfister(2015)]{bohm2015people}
Gisela B{\"o}hm and Hans-R{\"u}diger Pfister.
\newblock How people explain their own and others’ behavior: a theory of lay
  causal explanations.
\newblock \emph{Frontiers in psychology}, 6:\penalty0 139, 2015.

\bibitem[Boutilier et~al.(1995)Boutilier, Dearden, Goldszmidt,
  et~al.]{boutilier1995exploiting}
Craig Boutilier, Richard Dearden, Moises Goldszmidt, et~al.
\newblock Exploiting structure in policy construction.
\newblock In \emph{IJCAI}, volume~14, pages 1104--1113, 1995.

\bibitem[Braun and Clarke(2006)]{braun2006using}
Virginia Braun and Victoria Clarke.
\newblock Using thematic analysis in psychology.
\newblock \emph{Qualitative research in psychology}, 3\penalty0 (2):\penalty0
  77--101, 2006.

\bibitem[Brockman et~al.(2016)Brockman, Cheung, Pettersson, Schneider,
  Schulman, Tang, and Zaremba]{openai}
Greg Brockman, Vicki Cheung, Ludwig Pettersson, Jonas Schneider, John Schulman,
  Jie Tang, and Wojciech Zaremba.
\newblock {OpenAI Gym}, 2016.

\bibitem[Buhrmester et~al.(2011)Buhrmester, Kwang, and
  Gosling]{buhrmester2011amazon}
Michael Buhrmester, Tracy Kwang, and Samuel~D Gosling.
\newblock Amazon's mechanical turk: A new source of inexpensive, yet
  high-quality, data?
\newblock \emph{Perspectives on psychological science}, 6\penalty0
  (1):\penalty0 3--5, 2011.

\bibitem[Byrne(2019)]{byrne2019counterfactuals}
Ruth~MJ Byrne.
\newblock Counterfactuals in explainable artificial intelligence ({XAI}):
  evidence from human reasoning.
\newblock In \emph{Proceedings of the Twenty-Eighth International Joint
  Conference on Artificial Intelligence, IJCAI-19}, pages 6276--6282, 2019.

\bibitem[Chapman and Kaelbling(1991)]{chapman1991input}
David Chapman and Leslie~Pack Kaelbling.
\newblock Input generalization in delayed reinforcement learning: An algorithm
  and performance comparisons.
\newblock In \emph{IJCAI}, volume~91, pages 726--731. Citeseer, 1991.

\bibitem[De~Graaf and Malle(2017)]{de2017}
Maartje~MA De~Graaf and Bertram~F Malle.
\newblock How people explain action (and autonomous intelligent systems should
  too).
\newblock In \emph{2017 AAAI Fall Symposium Series}, 2017.

\bibitem[Elizalde and Sucar(2009)]{elizalde2009expert}
Francisco Elizalde and Luis~Enrique Sucar.
\newblock Expert evaluation of probabilistic explanations.
\newblock In \emph{ExaCt}, pages 1--12, 2009.

\bibitem[Elizalde et~al.(2007)Elizalde, Sucar, Reyes, and
  Debuen]{elizalde2007mdp}
Francisco Elizalde, Luis~Enrique Sucar, Alberto Reyes, and Pablo Debuen.
\newblock An {MDP} approach for explanation generation.
\newblock In \emph{ExaCt}, pages 28--33, 2007.

\bibitem[Elizalde et~al.(2008)Elizalde, Sucar, Luque, Diez, and
  Reyes]{elizalde2008policy}
Francisco Elizalde, L~Enrique Sucar, Manuel Luque, J~Diez, and Alberto Reyes.
\newblock Policy explanation in factored markov decision processes.
\newblock In \emph{Proceedings of the 4th European Workshop on Probabilistic
  Graphical Models (PGM 2008)}, pages 97--104, 2008.

\bibitem[Fukuchi et~al.(2017)Fukuchi, Osawa, Yamakawa, and
  Imai]{fukuchi2017autonomous}
Yosuke Fukuchi, Masahiko Osawa, Hiroshi Yamakawa, and Michita Imai.
\newblock Autonomous self-explanation of behavior for interactive reinforcement
  learning agents.
\newblock In \emph{Proceedings of the 5th International Conference on Human
  Agent Interaction}, pages 97--101. ACM, 2017.

\bibitem[Gunning and Aha(2019)]{gunning2019darpa}
David Gunning and David~W Aha.
\newblock Darpa's explainable artificial intelligence program.
\newblock \emph{AI Magazine}, 40\penalty0 (2):\penalty0 44--58, 2019.

\bibitem[Halpern and Pearl(2005)]{halpern2005causes}
Joseph~Y Halpern and Judea Pearl.
\newblock Causes and explanations: A structural-model approach. part {II}:
  Explanations.
\newblock \emph{The British journal for the philosophy of science}, 56\penalty0
  (4):\penalty0 889--911, 2005.

\bibitem[Hayes and Shah(2017)]{hayes2017improving}
Bradley Hayes and Julie~A Shah.
\newblock Improving robot controller transparency through autonomous policy
  explanation.
\newblock In \emph{Proceedings of the 2017 ACM/IEEE international conference on
  human-robot interaction}, pages 303--312. ACM, 2017.

\bibitem[Hilton(1990)]{hilton1990conversational}
Denis~J Hilton.
\newblock Conversational processes and causal explanation.
\newblock \emph{Psychological Bulletin}, 107\penalty0 (1):\penalty0 65, 1990.

\bibitem[Hilton et~al.(2005)Hilton, McClure, and Slugoski]{hilton2005course}
Denis~J Hilton, John~L McClure, and Ben~R. Slugoski.
\newblock The course of events: counterfactuals, causal sequences, and
  explanation.
\newblock In D.R. Mandel, D.J. Hilton, and P.~Catellani, editors, \emph{The
  {P}sychology of {C}ounterfactual {T}hinking}, pages 56--72. Routledge, 2005.

\bibitem[Hilton et~al.(2010)Hilton, McClure, and Sutton]{hilton2010selecting}
Denis~J Hilton, John McClure, and Robbie~M Sutton.
\newblock Selecting explanations from causal chains: Do statistical principles
  explain preferences for voluntary causes?
\newblock \emph{European Journal of Social Psychology}, 40\penalty0
  (3):\penalty0 383--400, 2010.

\bibitem[Hoffman et~al.(2018)Hoffman, Mueller, Klein, and
  Litman]{hoffman2018metrics}
Robert~R Hoffman, Shane~T Mueller, Gary Klein, and Jordan Litman.
\newblock Metrics for explainable {AI}: Challenges and prospects.
\newblock \emph{arXiv preprint arXiv:1812.04608}, 2018.

\bibitem[Hornsby(1993)]{hornsby1993agency}
Jennifer Hornsby.
\newblock Agency and causal explanation.
\newblock 1993.

\bibitem[Juozapaitis et~al.(2019)Juozapaitis, Koul, Fern, Erwig, and
  Doshi-Velez]{juozapaitis2019explainable}
Zoe Juozapaitis, Anurag Koul, Alan Fern, Martin Erwig, and Finale Doshi-Velez.
\newblock Explainable reinforcement learning via reward decomposition.
\newblock In \emph{Proceedings of the IJCAI 2019 Workshop on Explainable
  Artificial Intelligence}, pages 47--53, 2019.

\bibitem[Keren(2014)]{keren2014between}
Gideon Keren.
\newblock Between-or within-subjects design: A methodological dilemma.
\newblock \emph{A Handbook for Data Analysis in the Behaviorial Sciences},
  1:\penalty0 257--272, 2014.

\bibitem[Khan et~al.(2009)Khan, Poupart, and Black]{khan2009minimal}
Omar~Zia Khan, Pascal Poupart, and James~P Black.
\newblock Minimal sufficient explanations for factored markov decision
  processes.
\newblock In \emph{ICAPS}, 2009.

\bibitem[Klein(2018)]{klein2018explaining}
Gary Klein.
\newblock Explaining explanation, part 3: The causal landscape.
\newblock \emph{IEEE Intelligent Systems}, 33\penalty0 (2):\penalty0 83--88,
  2018.

\bibitem[Langley et~al.(2017)Langley, Meadows, Sridharan, and
  Choi]{langley2017explainable}
Pat Langley, Ben Meadows, Mohan Sridharan, and Dongkyu Choi.
\newblock Explainable agency for intelligent autonomous systems.
\newblock In \emph{Twenty-Ninth IAAI Conference}, 2017.

\bibitem[Lim et~al.(2009)Lim, Dey, and Avrahami]{lim2009and}
Brian~Y Lim, Anind~K Dey, and Daniel Avrahami.
\newblock Why and why not explanations improve the intelligibility of
  context-aware intelligent systems.
\newblock In \emph{Proceedings of the SIGCHI Conference on Human Factors in
  Computing Systems}, pages 2119--2128. ACM, 2009.

\bibitem[Lin(1992)]{lin1992self}
Long-Ji Lin.
\newblock Self-improving reactive agents based on reinforcement learning,
  planning and teaching.
\newblock \emph{Machine learning}, 8\penalty0 (3-4):\penalty0 293--321, 1992.

\bibitem[Lombrozo and Vasilyeva(2017)]{lombrozo2017causal}
Tania Lombrozo and Nadya Vasilyeva.
\newblock Causal explanation.
\newblock \emph{Oxford handbook of causal reasoning}, pages 415--432, 2017.

\bibitem[Madumal et~al.(2020)Madumal, Miller, Sonenberg, and
  Vetere]{madumal2019explainable}
Prashan Madumal, Tim Miller, Liz Sonenberg, and Frank Vetere.
\newblock Explainable reinforcement learning through a causal lens.
\newblock In \emph{Proceedings of the AAAI Conference on Artificial
  Intelligence}, 2020.

\bibitem[McClure and Hilton(1997)]{mcclure1997you}
John McClure and Denis Hilton.
\newblock For you can't always get what you want: When preconditions are better
  explanations than goals.
\newblock \emph{British Journal of Social Psychology}, 36\penalty0
  (2):\penalty0 223--240, 1997.

\bibitem[McClure et~al.(2007)McClure, Hilton, and Sutton]{mcclure2007judgments}
John McClure, Denis~J Hilton, and Robbie~M Sutton.
\newblock Judgments of voluntary and physical causes in causal chains:
  Probabilistic and social functionalist criteria for attributions.
\newblock \emph{European journal of social psychology}, 37\penalty0
  (5):\penalty0 879--901, 2007.

\bibitem[Miller(2018{\natexlab{a}})]{miller2018contrastive}
Tim Miller.
\newblock Contrastive explanation: A structural-model approach.
\newblock \emph{arXiv preprint arXiv:1811.03163}, 2018{\natexlab{a}}.

\bibitem[Miller(2018{\natexlab{b}})]{miller2018explanation}
Tim Miller.
\newblock Explanation in artificial intelligence: Insights from the social
  sciences.
\newblock \emph{Artificial Intelligence}, 2018{\natexlab{b}}.

\bibitem[Nagel and Stephan(2016)]{nagel2016explanations}
Jonas Nagel and Simon Stephan.
\newblock Explanations in causal chains: Selecting distal causes requires
  exportable mechanisms.
\newblock In \emph{Proceedings of the 38th Annual Conference of the Cognitive
  Science Society}, pages 806--812. Cognitive Science Society Austin, TX, 2016.

\bibitem[Roth et~al.(2019)Roth, Topin, Jamshidi, and
  Veloso]{roth2019conservative}
Aaron~M Roth, Nicholay Topin, Pooyan Jamshidi, and Manuela Veloso.
\newblock Conservative q-improvement: Reinforcement learning for an
  interpretable decision-tree policy.
\newblock \emph{arXiv preprint arXiv:1907.01180}, 2019.

\bibitem[Schuster and Paliwal(1997)]{schuster1997bidirectional}
Mike Schuster and Kuldip~K Paliwal.
\newblock Bidirectional recurrent neural networks.
\newblock \emph{IEEE Transactions on Signal Processing}, 45\penalty0
  (11):\penalty0 2673--2681, 1997.

\bibitem[Schwab and Karlen(2019)]{schwab2019cxplain}
Patrick Schwab and Walter Karlen.
\newblock Cxplain: Causal explanations for model interpretation under
  uncertainty.
\newblock In \emph{Advances in Neural Information Processing Systems}, pages
  10220--10230, 2019.

\bibitem[Sloman(2005)]{sloman2005causal}
Steven Sloman.
\newblock \emph{Causal models: How people think about the world and its
  alternatives}.
\newblock Oxford University Press, 2005.

\bibitem[Struckmeier et~al.(2019)Struckmeier, Racca, and
  Kyrki]{struckmeier2019autonomous}
Oliver Struckmeier, Mattia Racca, and Ville Kyrki.
\newblock Autonomous generation of robust and focused explanations for robot
  policies.
\newblock In \emph{2019 28th IEEE International Conference on Robot and Human
  Interactive Communication (RO-MAN)}, pages 1--8. IEEE, 2019.

\bibitem[Tabrez et~al.(2019)Tabrez, Agrawal, and Hayes]{tabrez2019explanation}
Aaquib Tabrez, Shivendra Agrawal, and Bradley Hayes.
\newblock Explanation-based reward coaching to improve human performance via
  reinforcement learning.
\newblock In \emph{2019 14th ACM/IEEE International Conference on Human-Robot
  Interaction (HRI)}, pages 249--257. IEEE, 2019.

\bibitem[Topin and Veloso(2019)]{topin2019generation}
Nicholay Topin and Manuela Veloso.
\newblock Generation of policy-level explanations for reinforcement learning.
\newblock In \emph{Proceedings of the AAAI Conference on Artificial
  Intelligence}, volume~33, pages 2514--2521, 2019.

\bibitem[van~der Waa et~al.(2018)van~der Waa, van Diggelen, Bosch, and
  Neerincx]{van2018contrastive}
Jasper van~der Waa, Jurriaan van Diggelen, Karel van~den Bosch, and Mark
  Neerincx.
\newblock Contrastive explanations for reinforcement learning in terms of
  expected consequences.
\newblock \emph{arXiv preprint arXiv:1807.08706}, 2018.

\bibitem[Vinyals et~al.(2017)Vinyals, Ewalds, Bartunov, Georgiev, Vezhnevets,
  Yeo, Makhzani, K{\"u}ttler, Agapiou, Schrittwieser,
  et~al.]{vinyals2017StarCraft}
Oriol Vinyals, Timo Ewalds, Sergey Bartunov, Petko Georgiev, Alexander~Sasha
  Vezhnevets, Michelle Yeo, Alireza Makhzani, Heinrich K{\"u}ttler, John
  Agapiou, Julian Schrittwieser, et~al.
\newblock Starcraft ii: A new challenge for reinforcement learning.
\newblock \emph{arXiv preprint arXiv:1708.04782}, 2017.

\bibitem[Wang et~al.(2019)Wang, Yang, Abdul, and Lim]{wang2019designing}
Danding Wang, Qian Yang, Ashraf Abdul, and Brian~Y Lim.
\newblock Designing theory-driven user-centric explainable ai.
\newblock In \emph{Proceedings of the 2019 CHI Conference on Human Factors in
  Computing Systems}, page 601. ACM, 2019.

\bibitem[Wang et~al.(2016)Wang, Pynadath, and Hill]{wang2016trust}
Ning Wang, David~V Pynadath, and Susan~G Hill.
\newblock Trust calibration within a human-robot team: Comparing automatically
  generated explanations.
\newblock In \emph{The Eleventh ACM/IEEE International Conference on Human
  Robot Interaction}, pages 109--116. IEEE Press, 2016.

\end{thebibliography}

\end{document}